%% file: main.tex
\def\isarxivversion{1} 

\ifdefined\isarxivversion
\documentclass[11pt]{article}
\usepackage[numbers]{natbib}
\else

\documentclass{article}
\usepackage{hyperref}
\usepackage{neurips_2023}

\fi

\usepackage[utf8]{inputenc}

\usepackage{amsmath}
\usepackage{amssymb}
\usepackage{amsthm}
\usepackage{mathtools}
\usepackage{bbm}
\usepackage{bm}
\usepackage{verbatim}
\usepackage{algorithm}
\usepackage{booktabs}
\usepackage{array}
\usepackage{setspace}
\usepackage{caption}
\usepackage{subcaption}
\usepackage{enumitem}
\usepackage{cases}
\usepackage{xcolor}
\usepackage{soul}

\usepackage{longtable}

\usepackage{algpseudocode}

\usepackage{amsfonts}

\ifdefined\isarxivversion
\usepackage{hyperref}
\hypersetup{colorlinks=true,citecolor=red,linkcolor=blue}
\usepackage[margin=1in]{geometry}
\else
\usepackage{hyperref}
\definecolor{mydarkblue}{rgb}{0,0.08,0.45}
\hypersetup{colorlinks=true, citecolor=mydarkblue,linkcolor=mydarkblue}
\fi

\newcommand{\OPT}{\mathrm{OPT}}
\newcommand{\R}{\mathbb{R}}

\renewcommand{\hat}{\widehat}
\renewcommand{\tilde}{\widetilde}

\DeclareMathOperator*{\argmax}{arg\,max}
\DeclareMathOperator*{\argmin}{arg\,min}
\DeclareMathOperator*{\E}{\mathbb{E}}

\newtheorem{theorem}{Theorem}[section]
\newtheorem{lemma}[theorem]{Lemma}
\newtheorem{definition}[theorem]{Definition}

\newtheorem{assumption}[theorem]{Assumption}

\newtheorem{fact}[theorem]{Fact}

\newtheorem{claim}[theorem]{Claim}

\begin{document}

\title{Clustered Linear Contextual Bandits with Knapsacks}

\ifdefined\isarxivversion
\author{
Yichuan Deng\thanks{\texttt{ycdeng@cs.washington.edu}. The University of Washington}
\and
Michalis Mamakos\thanks{\texttt{mamakosmm@gmail.com}. Northwestern University}
\and 
Zhao Song\thanks{\texttt{zsong@adobe.com}. Adobe Research}
}
\date{}
\else
\fi

\ifdefined\isarxivversion
\begin{titlepage}
  \maketitle
  \begin{abstract}

\input{abstract}

  \end{abstract}
  \thispagestyle{empty}
\end{titlepage}
{\hypersetup{linkcolor=black}
\tableofcontents
}
\newpage
\else
\maketitle
\begin{abstract}
\input{abstract}
\end{abstract}
\fi


\input{content}

\ifdefined\isarxivversion
\bibliographystyle{alpha}
\bibliography{bib_all}
\else
\bibliographystyle{alpha}
\bibliography{bib_all}
\fi
\onecolumn
\appendix
\section*{Appendix}

\input{appendix}

\end{document}

%% file: abstract.tex
In this work, we study clustered contextual bandits where rewards and resource consumption are the outcomes of cluster-specific linear models. The arms are divided in clusters, with the cluster memberships being unknown to an algorithm. Pulling an arm in a time period results in a reward and in consumption for each one of multiple resources, and with the total consumption of any resource exceeding a constraint implying the termination of the algorithm. Thus, maximizing the total reward requires learning not only models about the reward and the resource consumption, but also cluster memberships. We provide an algorithm that achieves regret sublinear in the number of time periods, without requiring access to all of the arms. In particular, we show that it suffices to perform clustering only once to a randomly selected subset of the arms. To achieve this result, we provide a sophisticated combination of techniques from the literature of econometrics and of bandits with constraints.

%% file: content.tex
\section{Introduction}

In the contextual bandits problem \cite{langford2007,slivkins2011,agarwal2014} a decision-maker sequentially, over a number of time periods, observes context that is informative about the period-specific reward that different choices (bandits) can provide. Nevertheless, in a time period, the actual reward is observed only for the choice made in that period. Therefore, learning over time the relation between the context and the reward of a choice requires, in some time periods, not making other choices about which such relation has already been adequately learned. This trade-off is commonly referred to in the literature of online learning as the exploration-exploitation dilemma \cite{sutton2018}, and it is encountered in a plethora of real-world problems that have  naturally been modeled as contextual bandits, including online recommender systems \cite{tang2014}, crowdsourcing \cite{abraham2013}, online advertising \cite{tang2013}, and applications in the health domain \cite{tewari2017}. Such problems often involve constraints about the choices of the decision-maker. For instance, an advertiser can be adapting over time the population segment that an ad campaign is targeted at \cite{schwartz2017}, but the total cost of the campaign can be subject to a budget. Thus, the successful formulation of a problem as a contextual bandits problem requires the incorporation of such constraints. Furthermore, the heterogeneity of the choices that are available to the decision-maker affects the performance of a contextual bandits algorithm. If the relation between the context and the reward is different for every choice, the exploration that an algorithm has to do increases with the number of the choices. This issue has motivated the consideration of linear contextual bandits \cite{chu2011} where the reward is specified by a linear model that is common among all the available choices. However, ignoring any systematic aspect of heterogeneity among the choices can imply unrealistic assumptions. A more realistic alternative is to consider the case where the available choices are clustered and with each cluster corresponding to a different linear model. For example, this can be the case when the effectiveness of an advertisement can vary across individuals but not significantly among those with similar demographics.

The need to accommodate the complexities involved in real-world applications has motivated a great number of recent works in the literature of multi-armed bandits. In one of the lines of such works, the constraints that a decision-maker faces are modeled as knapsacks \cite{agrawal2016a,badanidiyuru2018,immorlica2019}, with the arm played (choice made) in a time period resulting not only in a reward but also in the consumption of limited resources. Such knapsack constraints have given rise to the connection of bandit problems with the online stochastic packing problem \cite{feldman2010,devanur2011}. In a separate line of works, bandits problems with clusters are employed to model recommender systems  \cite{gentile2014,gentile2017,li2018a,ban2021}. In these works, at each time period a user arrives at the system seeking for recommendations, and with the users forming clusters based on their preferences. Then, the reward that the system receives in a  period depends on the recommendations it provides to the user. It is among our beliefs that  further problems can be modeled as bandits with clusters, and with cases where the bandits themselves form the clusters raising interesting questions, e.g., problems in online advertising where the users to see an ad must be chosen. Despite that constraints and clusters have been identified by the literature of online learning with bandits as important components of real-world applications, we are not aware of any previous work that takes both of them into account.

Thus, one natural question to ask is:
\begin{center}
    {\it Is it possible to give an algorithm with provable guarantees for the problem of contextual bandits with clusters and knapsacks?}
\end{center}
In this work, we provide a positive answer to the above question.

More specifically, we provide a contextual bandits algorithm with regret sublinear in the number of time periods, considering multiple resources and where the depletion of any resource implies the termination of the algorithm. Furthermore, the arms are clustered, with  cluster memberships being unknown to the algorithm. In each time period, the algorithm observes the i.i.d. context of each arm, and with the choice of an arm in a period resulting in a reward and the consumption of $d$ different resources. While the goal of the algorithm is to maximize the sum of the rewards over periods, if the accumulated consumption of any of the $d$ resources exceeds a given budget then the algorithm must terminate. Moreover, the reward and consumption distributions of an arm are specified by cluster-specific functions that are linear in the context. Our model extends that of \cite{agrawal2016a} by considering clusters. The approach we employ builds on the algorithm of that paper, but the main challenge we face is that arms have to be clustered while ensuring that the average regret is still vanishing in the number of periods.

Our algorithm needs to perform clustering only once. Initially, a subset of the total $K$ arms is sampled, with this subset being denoted by $\mathcal{S}$, and with arms not in $\mathcal{S}$ never being played. The main intuition behind this first step is that to derive vanishing average regret in the number of periods, we need to have access to arms of any of the clusters and to be able to accurately estimate the proportion of the total $K$ arms that a cluster consists of.  Once $\mathcal{S}$ is sampled, every arm in $\mathcal{S}$ is played a specified number of times in order to collect sufficiently many observations to achieve accurate clustering. Notice that playing an arm gives $d+1$ observations, one for the reward and $d$ for the consumed resources, with each of them coming from a different linear model. Then, for the clustering of the arms in $\mathcal{S}$ we borrow the classifier-Lasso method from the literature of econometrics \cite{su2016}. This method treats the parameters of the arms as  distinct, but not necessarily unequal, to the parameters of the clusters. In the time periods after the clustering of the arms in $\mathcal{S}$, the algorithm chooses the arm to play by following the principle of optimism in the face of uncertainty \cite{auer2002a,auer2002b,abbasi2011}. Thus, in every period an optimistic estimate about the reward and the consumption of each arm in $\mathcal{S}$ is made that depends on the observed context and the estimated linear models of each cluster. However, since the clustering may not be perfect, some of the arms may be wrongly grouped with arms that do not share the same true cluster with them. Consequently, not all of the arms assigned to a cluster necessarily share the same true parameters. From a technical perspective, in order to derive formal results about the regret, we treat all of the arms assigned to a cluster as belonging to the same true cluster, but where the error term of the linear models, with some probability that depends on the error of the clustering, has expected value that depends on the context instead of  zero. In other words, we account for the error due to imperfect clustering by treating it as measurement error in the context \cite{wansbeek2001, fuller2009}. Moreover, known results from the literature of convex online learning \cite{srebro2011} are exploited in order to take into account the amount that has been depleted from each of the $d$ resources, as suggested by the algorithm proposed in \cite{agrawal2016a}.

In regard to the number of time periods $T$, the regret of our algorithm increases at the rate $T^{1 - \delta}$, for $\delta \in (0, \frac{1}{2})$, requiring the budget for each resource to be $B > \tilde{O}(T^{2 \delta})$.  
Thus, the regret is sublinear in $T$, and it can  approach the square root rate as the resource budget approaches $T$. However, as it becomes clear in the next section, we must have $B < T$, since otherwise we could ignore the constraints. The main drawback of our algorithm is that its regret depends on the number of arms $K$, as it operates only on a subset of them. Nevertheless, our results suggest that such dependence may be inevitable. Overall, we believe that our approach points to a fruitful direction for the consideration of online learning problems that involve clusters and constraints.

\subsection{Related Work}
Some of the earliest works to consider the problem of bandits with knapsacks were \cite{badanidiyuru2014} and an earlier version of \cite{badanidiyuru2018}, with both of them considering stochastic reward and resource consumption. The adversarial version of the problem was initially considered in \cite{immorlica2019}, and consequently in \cite{kesselheim2020}. An algorithm that achieves low expected regret for both the adversarial and the stochastic version is investigated in \cite{rangi2019}, while in \cite{amani2019} bandit algorithms with constraints for safety-critical systems are proposed. General forms for the correlation of the reward and the resource consumption are considered in \cite{cayci2020}. \cite{devanur2011}, \cite{agrawal2014}, and \cite{agrawal2016b} study computationally efficient algorithms for online learning problems with constraints, including bandits with knapsacks. In a problem similar to that considered by \cite{agrawal2016a}, \cite{wu2015} considers the case where there is a single resource. From a result of \cite{dani2008}, it is known that the regret of a linear contextual bandits algorithm must be at least linear in the dimensionality of the context. The topic of contextual bandits with behavioral constraints is presented in \cite{balakrishnan2018}, while \cite{yang2020} studies contextual bandits with a resource constraint in recommender systems \cite{carlsson2021}. 

In works that model recommender systems as bandits with clusters of users \cite{nguyen2014,gentile2014,gentile2017,li2018a,ban2021,li2021}, the users arrive at the system exogenously. Besides the consideration of constraints and the fact that the model we study is not specific to an application, we cannot adopt techniques from the literature of recommender systems because we consider clusters of arms. Thus, the units to perform clustering on are endogenously and not exogenously provided, which makes our problem harder. 

Clustering under mixed linear models (like ours) has attracted a lot of the interest in the literature of computer science \cite{zhong2016,li2018b,chen2020,kong2020,chen2021}, but with the results of these works typically requiring a lower bound on the number of units to be clustered. Interestingly, similar versions of this problem have been considered in the literature of econometrics \cite{lin2012,ando2016,su2016,su2019,gu2019,okui2021}. Here, we exploit results from the latter literature, since the assumptions they require are more appropriate to our case. In particular, we make assumptions about the joint rate of the number of sampled arms and of the observations collected for each such arm, instead of imposing a lower bound on the former.

\subsection{Organization}
The rest of the paper is organized as follows. In Section~\ref{sec:problem}, we provide notation and define the problem under consideration. Technical ingredients that are utilized throughout the paper are presented in Section~\ref{sec:preliminary}. Section~\ref{sec:uncertainty} is devoted to presenting how the principle of optimism in the face of uncertainty is employed. The algorithm and results about its regret are presented in Section~\ref{sec:algorithm}. Finally, in Section~\ref{sec:discussion} we conclude with a discussion about our findings and future directions. Proofs are deferred to the appendix.

\section{Problem Definition and Notation}\label{sec:problem}
In this section, we first introduce notation that we use throughout the paper, and then we provide the definition of the problem we consider. For any positive integer $N$, we use $[N]$ to denote the set $\{1,2,\cdots,N\}$. We use $\E[X]$ to denote the expectation of the random variable $X$, and $\Pr[E]$ to denote the probability of the event $E$. The notation $\mathbbm{1}[f]$ is used for the indicator variable which outputs $1$ if $f$ holds and $0$ otherwise. Boldface lower case letters are reserved for vectors, and boldface upper case letters for matrices. The vector $\bm{1}_d$ denotes the $d$-dimensional vector that has the value $1$ in every dimension, and  $\bm{0}_d$ is defined equivalently. Calligraphic upper case letters, e.g., $\mathcal{S}$, denote sets. The probability simplex in $d$ dimensions is denoted as $\Delta^d$. For a square matrix $\bm{A}$, we define the matrix norm of the vector $\bm{x}$ as $\| \bm{x} \|_{\bm{A}} = \sqrt{\bm{x}^\top \bm{A} \bm{x}}$.

The number of arms is denoted by $K \in \mathbb{N}$, and the number of clusters by $C \in \mathbb{N}$. Each arm belongs to one cluster, with $c(a) \in [C]$ denoting the cluster of arm $a \in [K]$. The algorithm is given a budget $B \in \mathbb{R}_+$ and the number of clusters $C$, but not the arm membership in the clusters. In each period $t \in [T]$, the algorithm observes the context of all the arms, with $\bm{x}_t(a) \in [0, 1]^m$ denoting the context of arm  $a$ in period $t$, then chooses arm $a_t \in [K]$, and finally observes reward $r_t(a_t) \in [0, 1]$ and consumption vector $\bm{v}_t(a_t) \in [0, 1]^d$. The algorithm can also play the ``no-op'' action that deterministically gives zero reward and consumption. The goal of the algorithm is to maximize the total reward $\sum_{t \in [T]} r_t(a_t)$ under the constraints $\sum_{t \in [T]} \bm{v}_t(a_t) \leq B  \cdot \bm{1}_d$. Notice that it is without loss of generality to consider the same budget for each of the $d$ resources, since this can imposed by normalization.

Moreover, we assume that the reward and the consumption are generated by cluster-specific linear models, as indicated by the assumption below.

\begin{assumption}[Linearity]
\label{assumption:linearity}
For each cluster $c \in [C]$ there is an unknown vector $\bm{\mu}_c \in [0,1]^m$ and an unknown matrix $\bm{W}_c \in [0,1]^{m \times d}$ such that for error terms $g_t(a)$ and $\bm{q}_t(a)$, with $a \in [K]$, it holds that
\begin{align}
        & r_t(a) = \bm{\mu}_{c(a)}^\top \bm{x}_t(a) + g_t(a)
        \label{eq:reward-model}
        \\
        & \bm{v}_t(a)  = \bm{W}_{c(a)}^\top \bm{x}_t(a) + \bm{q}_t(a)
        \label{eq:consumption-model}
\end{align}

where $\bm{\mu}_{c(a)} = \bm{\mu}_c$ and $\bm{W}_{c(a)} = \bm{W}_c$  if $c(a) = c$.
\end{assumption}

We make the following assumptions about the error terms.

\begin{assumption}[Zero conditional mean]
$\E[g_t(a) | \bm{x}_t(a)] = 0 \text{ and } \E[\bm{q}_t(a) | \bm{x}_t(a)] = \bm{0}_d$.
\label{assumption:zero-mean}
\end{assumption}

\begin{assumption}
$|g_t(a)| \leq 2R$ and $\| \bm{q}_{t}(a) \|_\infty \leq 2R$.
\end{assumption}

Furthermore, we make the following i.i.d. assumption about the context.

\begin{assumption}[i.i.d.]
The context $\bm{x}_t(a)$ is i.i.d. across arms and periods.
\label{assumption:iid}
\end{assumption}

We let the vector $\bm{p} = (p_1, \ldots, p_C) \in \Delta^C$ denote the proportions of arms that are in each cluster, i.e., $p_c = \frac{1}{K}\sum_{a \in [K]} \mathbbm{1}[c(a) = c]$ for $c \in [C]$. The smallest proportion in $\bm{p}$ is denoted as
$
    p_{\min} := \min_{c \in [C]} p_c
$. 
Even though the algorithm does not know the other proportions in $\bm{p}$, some knowledge about $p_{\min}$ is required for the clustering, as described later. Also, notice that $p_{\min} \leq 1/C$ by definition.

\subsection{Benchmark}
Our goal is to devise an algorithm that achieves sublinear regret in the number of time periods $T$. We employ as benchmark the expected reward of the optimal static policy which needs to satisfy the consumption constraints only in expectation. This benchmark policy knows the reward and consumption parameters for each cluster, as well as the cluster membership of each arm. It is known \cite{devanur2011, agrawal2016a, badanidiyuru2018} that the optimal static policy achieves the same expected reward to the optimal adaptive policy which knows the distribution of the context and needs to satisfy the constraints for the realizations of the resource consumption.

\begin{definition}[Optimal Static Policy]
Let $\bm{X} = \big( \bm{x}(1), \ldots, \bm{x}(K) \big) \in [0, 1]^{m \times K}$ be the matrix with the context of all arms in an arbitrary period, and $\pi(i, \bm{X}) \in [0, 1]$ the probability with which the action $i \in [K] \cup \{ \text{``no-op''} \}$ is taken by the static policy $\pi$ when the context is $\bm{X}$. The per-period expected reward and consumption vector of $\pi$ are respectively defined as
\begin{align}\label{eq:reward-function}
r(\pi) := & ~ \E_{\bm{X}}\Big[  \sum_{a \in [K]} \bm{\mu}_{c(a)}^\top \bm{x}(a) \pi(a, \bm{X}) \Big] , \notag \\
\bm{v}(\pi) := & ~ \E_{\bm{X}}\Big[  \sum_{a \in [K]} \bm{W}_{c(a)}^\top \bm{x}(a) \pi(a, \bm{X}) \Big]
\end{align}
With $\Pi$ denoting the set of all static policies, the optimal static policy is defined as
\begin{align*}
    \pi^* :=  \argmax_{\pi \in \Pi} r(\pi) \text{ subject to } \bm{v}(\pi) \leq \frac{B}{T} \cdot \bm{1}_d .
\end{align*}
The expected total reward of $\pi^*$ is defined as
\begin{align*}
    \mathrm{OPT} := T \cdot r(\pi^*) .
\end{align*}
\end{definition}

Since the ``no-op'' action is allowed, the policy $\pi^*$ is feasible and the definition of $\OPT$ is valid. Having specified the benchmark, we can now define the regret of our algorithm. 

\begin{definition}[Regret]
The regret of the algorithm for $T$ time periods is defined as
\begin{align}
    \mathrm{regret} (T) := \mathrm{OPT} - \sum_{t=1}^T r_t(a_t)
    \label{eq:def-regret}
\end{align}

\end{definition}

\section{Preliminaries}\label{sec:preliminary}
In this section we present results and technical ingredients that will allow us to derive the regret.


\subsection{Confidence Ellipsoid}
\label{subsection:ce}
In this subsection we present results about bounds on the parameter estimates that will be derived by the algorithm after the clustering is conducted. The approach here is to express the clustering error as violation of the zero conditional mean assumption about the error terms. Then, the technique of confidence ellipsoids, which is standard in the literature of online learning, is employed to bound the parameter estimates. The definition that follows considers a linear model for a response variable $y_t$ that generalizes the response variables in the problem we consider, i.e., $r_t(a)$ and each dimension of $\bm{v}_t(a)$. Therefore, the results derived in this subsection will imply results about our actual problem.

\begin{definition}\label{def:y_t}
We define $y_t \in [0, 1]$ as
    $   y_t := \bm{\mu}^\top \bm{x}_t + \eta_t$
where $\bm{\mu}, \bm{x}_t \in [0, 1]^m$, with $\bm{\mu}^\top \bm{x}_t \in [0, 1]$, and $\eta_t$  an error term that is the sum of two random  variables,
 $  \eta_t = u_t + h_t$
where
  $  \E[u_t|\bm{x}_t] = 0, |u_t| \leq 2R$
and for $\bm{\gamma} \in [-1, 1]^m$,
\begin{align*}
  h_t = 
  \begin{cases}
    0 & \text{w.p. } 1 - \epsilon \\
    \bm{\gamma}^\top \bm{x}_t & \text{w.p. } \epsilon 
  \end{cases}
\end{align*}
 \end{definition}
 
This definition corresponds to a measurement error model \cite{wansbeek2001,fuller2009}, as the expected value of the error term is $\E[\eta_t | \bm{x}_t] = \epsilon \bm{\gamma}^\top \bm{x}_t$, which in the general case is not  zero. The error term $u_t$ corresponds to an error term of the model of our actual problem, i.e., $u_t$ represents $g_t(a)$ and the individual dimensions of $\bm{q}_t(a)$. Since the context is i.i.d. across arms and periods, the probability $\epsilon$ is perceived as the clustering error in Eq. (\ref{eq:clustering-error}). Thus, $\bm{\gamma}$ is perceived as the element-wise difference between parameters of different clusters. For instance, considering $C=2$ and $y_t$ corresponding to $r_t(a)$ for $c(a) = 1$, we have that $\bm{\gamma}$ corresponds to $\bm{\mu}_2 - \bm{\mu}_1$. Therefore, $h_t$ represents the part of the error term due to $\bm{x}_t$ being endogenous, when zero clustering error is falsely assumed. We let $h_t$ have two branches instead of $C$ because $\bm{\gamma}$ can be perceived as a bound on parameter differences.

\begin{definition}
For regularization parameter $\lambda_2 > 0$, let
\begin{align}\label{eq:ols}
    \bm{M}_t := & ~ \lambda_2 \bm{I}_m + \sum_{i=1}^{t-1} \bm{x}_i \bm{x}_i^\top , \notag \\
    \hat{\bm{\mu}}_t := & ~ \bm{M}_t^{-1} \sum_{i=1}^{t-1} \bm{x_i} y_i 
\end{align}

\end{definition}
 
  \begin{definition} For any $\zeta \in (0, 1)$, define the confidence ellipsoid at time $t$ as
 \begin{align*}
    ~ & \mathcal{C}_t := \Big\{ \bm{\beta} \in \R^m : \| \bm{\beta} - \hat{\bm{\mu}}_t \|_{\bm{M}_t} \leq \\
    ~ &  2(R+1) \sqrt{ m \log (  t m /(\lambda_2\zeta) ) } + \epsilon m \sqrt{t} + \sqrt{\lambda_2 m}  \Big\}
 \end{align*}
 
 \end{definition}

The next two lemmas are the results derived in this subsection.

\begin{lemma}
 With probability at least $1 - \zeta$, $\bm{\mu} \in \mathcal{C}_t$.
 \label{lemma:ellipsoid}
\end{lemma}
 
 \begin{lemma}[Sum of rewards]\label{lemma:sum-rewards}
Consider $\bm{\tilde{\mu}}_t \in  \mathcal{C}_t$. For $R = \frac{1}{2}$ and $\lambda_2 = 1$, with probability at least $1 - \zeta$
\begin{align*}
     & ~ \sum_{t=1}^T | \bm{\tilde{\mu}}_t^\top \bm{x}_t - \bm{\mu}^\top \bm{x}_t | \\
    \leq & ~ 4 m \sqrt{ T \log (  T m /{\zeta} ) \log(T)  } + \epsilon  m^{\frac{3}{2}} T \sqrt{\log(T)} .
\end{align*}

\end{lemma}

Lemma \ref{lemma:ellipsoid} implies that a parameter estimate lies within some fixed distance from the true parameter with some specified probability, while Lemma \ref{lemma:sum-rewards} serves as the basis for employing optimistic parameter estimates, as described later.

\subsection{Online Learning}
\label{subsection:online-learning}
A special case of the online convex optimization problem is the game where at period $t \in [T]$ a learner chooses 
\begin{align*} 
\bm{\theta}_t \in \{ \bm{\theta} \in [0, 1]^d :  \| \bm{\theta} \|_1 \leq 1 \}
\end{align*}
based on past observations, and the adversary chooses the learner's payoff to be the outcome of a function that is linear in $\bm{\theta}_t$. It is known \cite{srebro2011,shalev2011} that for $\bm{\theta}_t$ chosen based on the online mirror descent  algorithm, the regret against the best fixed action of the learner in the online convex optimization problem is $O(\sqrt{\log(d) T})$. In the context of our problem, we utilize this result following the approach of \cite{agrawal2016a}. In particular, $\bm{\theta}_t$ will allow the consideration of resources in the arm choices in periods $t > N_\mathcal{S} T_0$, 
decreasing thus the probability that a choice will lead to depletion of one of the $d$ resources and consequently to the termination of the algorithm. To derive the regret for our problem (Eq. (\ref{eq:def-regret})), we will consider as the payoff chosen by the adversary the value 
\begin{align*} 
\bm{\theta}_t^\top (\bm{v}_t(a_t) - \frac{B - N_\mathcal{S} T_0}{T - N_\mathcal{S} T_0} \bm{1}_d),
\end{align*}
as illustrated by the algorithm presented later.

\section{Optimism in the Face of Uncertainty}\label{sec:uncertainty}
In this section, we describe how optimistic estimates of the reward and consumption parameters are derived in time periods after the clustering is performed. Such estimates will consequently be utilized to devise a no-regret algorithm in the next section. At  period $t > N_{\mathcal{S}} \cdot T_0$, the parameters of cluster $c \in [C]$ are estimated using all the observations prior to $t$ that involve arms that have been assigned to $c$. We denote by $t_c < t$ the number of periods in which arms estimated to be in $c$ have been played prior to period $t$, i.e.,
$
    t_c := \sum_{i=1}^{t-1} \mathbbm{1}[\hat{c}(a_i) = c]
$.  
Considering $r_t(a)$, $\bm{\mu}_{c}$, and $\epsilon_{c}$ in the place of $y_t$, $\bm{\mu}$, and $\epsilon$ from Definition \ref{def:y_t}, respectively, for $c=\hat{c}(a)$, $a \in \mathcal{S}$, we can apply the results from subsection \ref{subsection:ce} to the reward parameter vector of cluster $c$ by taking into consideration the $t_c$ periods corresponding to this cluster. Denoting by $\bm{w}_{c, j}$ the $j^{\text{th}}$ column of $\bm{W}_c$, $j \in [d]$, we can equivalently apply the results from subsection \ref{subsection:ce} to the consumption parameters. Letting $R = \frac{1}{2}$, $\lambda_2 = 1$, and  
\begin{align}
     \bm{M}_{c, t} := \bm{I}_m + \sum_{i < t: \hat{c}(a_{i}) = c} \bm{x}_i(a_i) \bm{x}_i(a_i)^\top
     \label{eq:Mct}
\end{align}
\noindent the parameters of cluster $c \in [C]$ are estimated at period $t$ as
\begin{align}
    \hat{\bm{\mu}}_{c, t} &:= \bm{M}_{c, t}^{-1} \sum_{i < t: \hat{c}(a_{i}) = c} \bm{x}_i(a_i) r_i(a_i)
    \label{eq:est-mu}
    \\
    \hat{\bm{W}}_{c, t} &:= \bm{M}_{c, t}^{-1} \sum_{i < t: \hat{c}(a_{i}) = c} \bm{x}_i(a_i) \bm{v}_i(a_i)^\top
    \label{eq:est-W}
\end{align}
Now, we can define confidence ellipsoids for the parameters of each cluster and derive optimistic estimates. The confidence ellipsoid of the reward vector of cluster $c$ at period $t$ is defined as 
\begin{align*}
    & ~ \mathcal{C}_{\mu, c, t} := \{ \bm{\beta} \in \R^m : \| \bm{\beta} - \hat{\bm{\mu}}_{c,t} \|_{\bm{M}_{c,t}} \\
    \leq & ~ 3 \sqrt{ m \log ( t_c m /\zeta ) } + \epsilon m \sqrt{t_c} + \sqrt{m} \},
\end{align*} 
and the optimistic estimate of the reward parameter for arm $a \in S$ at period $t$ is defined as
\begin{align}
    \tilde{\bm{\mu}}_{a, t} := \argmax_{\bm{\beta}  \in \mathcal{C}_{\mu, \hat{c}(a), t}}  \bm{x}_t(a)^\top  \bm{\beta}
    \label{eq:optimistic-mu}
\end{align}

We can similarly define the confidence ellipsoid for the vector of the consumption dimension $j \in [d]$ for cluster $c$  at period $t$ as
\begin{align*}
     & ~ \mathcal{C}_{w, c, t, j} := \Big\{ \bm{\beta} \in \R^m : \| \bm{\beta} - \hat{\bm{w}}_{c,t,j} \|_{\bm{M}_{c,t}} \\
    \leq & ~ 3 \sqrt{ m \log (  d t_c m /\zeta ) } + \epsilon m \sqrt{t_c} + \sqrt{m} \Big\},
\end{align*}
where $\hat{\bm{w}}_{c,t,j}$ is the $j^\text{th}$ column of $\hat{\bm{W}}_{c,t}$. Given a vector $\bm{\theta}_t \in [0, 1]^d$, the optimistic estimate of the consumption matrix for arm $a \in \mathcal{S}$ at period $t$ is defined by taking into consideration the Cartesian product of the $d$ confidence ellipsoids about consumption,
\begin{align}
    \tilde{\bm{W}}_{a, t} := \argmin_{\bm{W} \in  \bigtimes_{j=1}^d \mathcal{C}_{w,\hat{c}(a),t,j}} \bm{x}_t(a)^\top \bm{W} \bm{\theta}_t
     \label{eq:optimistic-W}
\end{align}
By Lemma \ref{lemma:ellipsoid} and the union bound, $\bm{W}_{c}$ is in $\bigtimes_{j=1}^d \mathcal{C}_{w,c,t,j}$ with probability at least $1 - \zeta$. The vector $\bm{\theta}_t$ will allow the algorithm in the next section to translate consumption into reward, so that arms that are expected to consume plenty of scarce resources will be relatively less appealing. For the optimistic estimate of a reward vector, the maximizer is picked in Eq. (\ref{eq:optimistic-mu}) since we want the choice of an arm to result in high reward. However, since large budget losses are undesirable, the optimistic estimate of a consumption matrix is taken to be the minimizer in Eq. (\ref{eq:optimistic-W}). The following lemma relates the optimistic estimates to the true parameters.

\begin{lemma}
Given clustering $\{\hat{c}(a)\}_{a \in \mathcal{S}}$ and vectors $\{\bm{\theta}_i\}_{i=N_{\mathcal{S}} \cdot T_0 + 1}^t$, 
where $\bm{\theta}_i \in [0, 1]^d$, with probability at least $1 - \zeta$ we have that for any $a \in \mathcal{S}$,
\begin{itemize}
    \item[a)] $\bm{x}_t(a)^\top \big( \tilde{\bm{\mu}}_{a, t} -  \bm{\mu}_{\hat{c}(a)} \big) \geq 0$
    \item[b)] $\bm{x}_t(a)^\top \big( \tilde{\bm{W}}_{a, t} - \bm{W}_{\hat{c}(a)} \big) \bm{\theta}_t \leq 0$
    \item[c)]  
    \begin{align*} 
    & ~ |\sum_{i=N_{\mathcal{S}} \cdot T_0 + 1}^t  \bm{x}_i(a_i)^\top ( \tilde{\bm{\mu}}_{a_i, i}  -  {\bm{\mu}}_{\hat{c}(a_i)} ) | \\
    \leq & ~ 4 C m \sqrt{t \log \big( t m / \zeta \big) \log(t)} + \epsilon_c  m^\frac{3}{2} t \sqrt{\log(t)}
    \end{align*}
    \item[d)] 
    \begin{align*} 
    & ~ \| \sum_{i=N_{\mathcal{S}} \cdot T_0 + 1}^t  \bm{x}_i(a_i)^\top \big( \tilde{\bm{W}}_{a_i, i} - \bm{W}_{\hat{c}(a_i)}\big) \|_\infty \\ 
    \leq & ~ 4 C m \sqrt{t \log \big( t m / \zeta \big) \log(t)} + \epsilon_c  m^\frac{3}{2} t \sqrt{\log(t)}
    \end{align*}
\end{itemize}
\label{lemma:optimism}
\end{lemma}

\section{Algorithm}\label{sec:algorithm}
In this section, we present our algorithm for the problem of clustered linear contextual bandits with knapsacks (clusterLCBwK) and results related to its regret. Initially, clustering is performed as specified in subsection \ref{subsection:clustering}. Then, using the initial $N_\mathcal{S} T_0$ samples, the reward of the optimal static policy is estimated, as specified in the following lemma.

\begin{lemma}
\label{lemma:Z}
Let 
$\hat{\OPT}:= T \cdot \hat{r}$, 
where $\hat{r}$  is the estimate of $r(\pi^*)$ based on the initial $\mathcal{N_\mathcal{S}} T_0$ random samples, defined as
\begin{align*}
    & \hat{r} :=   \max_{\pi}  \frac{K}{N^2_{\mathcal{S}}  T_0} \sum_{t=1}^{N_{\mathcal{S}}  T_0}  \sum_{a \in \mathcal{S}} \hat{\bm{\mu}}_{\hat{c}(a), N_{\mathcal{S}}  T_0+1}^\top \bm{x}_t(a) \pi(a, \bm{X}_t) 
\\
& \text{s.t. }
     \frac{K}{N^2_{\mathcal{S}}  T_0} \sum_{t=1}^{N_{\mathcal{S}}  T_0}  \sum_{a \in \mathcal{S}} \hat{\bm{W}}_{\hat{c}(a), N_{\mathcal{S}}  T_0+1}^\top \bm{x}_t(a) \pi(a, \bm{X}_t) \\
     & \leq \frac{B}{T} \bm{1}_d
\end{align*}
Then, 
$
\hat{\OPT} - \OPT = o(1)
$
with high probability. 
\end{lemma}

The estimate $\hat{\OPT}$ is then used to define the variable $Z$ which will contribute to the choice made by the algorithm in periods $t>N_\mathcal{S} T_0$ by appropriately weighting the optimistic estimates of the consumption. In particular, this variable is defined as
\begin{equation}
    Z := \frac{N_\mathcal{S} \hat{\OPT}}{2 K B^\prime }
    \label{eq:Z}
\end{equation}
where $  B^\prime :=  B - N_\mathcal{S} T_0
$ and we also define $T^\prime$ similarly, i.e., 
$
T^\prime := T - N_\mathcal{S} T_0.
$
In periods $t > N_\mathcal{S} T_0$, the choice of the algorithm is 
\begin{align*} 
a_t = \argmax_{a \in \mathcal{S}} \bm{x}_t(a)^\top ( \tilde{\bm{\mu}}_{a, t} -  Z \tilde{\bm{W}}_{a, t} \bm{\theta}_t) ,
\end{align*}
where $\bm{\theta}_t \in [0, 1]^d$ is the choice of the online mirror descent algorithm when at the previous period the payoff
$
\bm{\theta}_{t-1}^\top (\bm{v}(a_{t-1}) - \frac{B^\prime}{T^\prime} \bm{1}_d)
$
is observed. Thus, in effect, $Z$ and $\bm{\theta}_t$ allow the resource consumption to be compared to the reward, so that arms estimated to consume a lot of a scarce resource can be avoided.

Our algorithm is an extension of the linear contextual bandits with knapsacks (linCBwK) algorithm \cite{agrawal2016a}, but with the clustering step having been incorporated and accounted for in the derivation of the regret. Moreover, our estimation of $\OPT$ utilizes the  initial randomly collected $N_{\mathcal{S}} T_0$  samples, and is different to the corresponding estimation in linCBwK which would lead to additional sampling and thus higher regret in our case.

\begin{algorithm}
  \caption{clusterLCBwK}
  \begin{algorithmic}[1]
    \State $N_{\mathcal{S}} \gets O \big( p^{-1}_{\min} ( T^\delta + \log C) \big)$
    \State $\mathcal{S} \gets$ random subset of $[K]$ with size $N_{\mathcal{S}}$
    \State $T_0 \gets N_{\mathcal{S}}$
    \State $\forall a \in \mathcal{S}$, play $T_0$ times the arm $a$ 
    \State Cluster the arms in $\mathcal{S}$ per Eq. (\ref{eq:clustering-method})
    \State Compute $Z$ per Eq. (\ref{eq:Z})
    \For{$t = N_{\mathcal{S}}  T_0 + 1, \ldots, T$}
        \State $\forall a \in \mathcal{S}$, obtain $\tilde{\bm{\mu}}_{a, t}$ and $\tilde{\bm{W}}_{a, t}$ per Eq. (\ref{eq:optimistic-mu}), (\ref{eq:optimistic-W})
        \State $a_t \gets \argmax_{a \in \mathcal{S}} \bm{x}_t(a)^\top ( \tilde{\bm{\mu}}_{a, t} -  Z \tilde{\bm{W}}_{a, t} \bm{\theta}_t)$
        \State Play the arm $a_t$, and observe $r_t(a_t)$ and $\bm{v}_t(a_t)$
        \State \textbf{if} $\exists j \in [d]:   \sum_{i=1}^t \bm{v}_{i}(a_i)^\intercal \bm{e}_j \geq B$ \textbf{then} \text{exit}
        \State Update $\bm{M}_{c, t+1}$, $\hat{\bm{\mu}}_{c,t+1}$, and $\hat{\bm{W}}_{c, t+1}$, for $c = \hat{c}(a_t)$
        \State Update $\bm{\theta}_{t+1}$ with the online mirror descent algorithm for payoff $\bm{\theta}_t^\top (\bm{v}_t(a_t) - \frac{B^\prime}{T^\prime} \bm{1}_d)$
    \EndFor
  \end{algorithmic}
  \label{alg:clusterLCBwK}
\end{algorithm}

The following theorem is our main result, showing that the regret of Algorithm \ref{alg:clusterLCBwK} is sublinear in $T$.

\begin{theorem}\label{theorem:main}
For $\delta \in (0, \frac{1}{2})$, 
and $B > N_\mathcal{S} T_0$, 
with high probability
\begin{align*}
     \mathrm{regret}(T) 
    \leq
     O \Big( R(T) \big( 1 + \frac{N_\mathcal{S} \OPT}{ K B^\prime} \big) +  \OPT \big( 1 -  \frac{N_\mathcal{S}}{K} \big) \Big)
\end{align*}
where 
\begin{align*}
 R(T) = O\Big(C  p_{\min}^{-1} m^\frac{3}{2} T^{1 - \delta} \sqrt{\log(T)} \Big).
\end{align*}
\end{theorem}


Notice that since 
$
\frac{N_\mathcal{S}}{K} \leq 1,
$
the exponent of $T$ is at most $1 - \delta < 1$, and thus the regret is sublinear in the number of time periods. Considering the case where all of the arms are initially sampled, i.e., $N_\mathcal{S} = K$, the bound on the regret is 
\begin{align*}
O\big( R(T) (1 + \frac{\OPT}{B^\prime}) \big),
\end{align*}
and we can compare it to the regret for the case without clusters,  which  is 
\begin{align*} 
O\big( R^\prime(T) (1 + \frac{\OPT}{B}) \big)
\end{align*}
\cite{agrawal2016a}, for
$
R^\prime(T) = \tilde{O}(m T^{\frac{1}{2}})
$
. The term $R(T)$ occurs because of errors in the clustering of the arms, while $R'(T)$ comes from errors in the estimation of the linear model parameters. While our algorithm also suffers $R'(T)$, it is dominated by $R(T)$. Although it would be interesting to consider whether repeating the clustering could reduce $R(T)$, the term $R'(T)$ would still persist. Moreover, the difference between 
\begin{align*} 
B^\prime = B - N_\mathcal{S} T_0
\end{align*}
and $B$ in the division of $\OPT$ reflects the loss due to the initial $N_\mathcal{S} T_0$ periods of obtaining samples for the clustering. The probability $1 - \zeta$ that appears in Lemma \ref{lemma:optimism} affects our regret, but only through $R'(T)$. Although $C$ is fixed by Assumption \ref{assumption:fixed-C}, we have kept in the presentation of the regret for transparency.

Considering the case that is of greater interest here, where a subset of the arms is sampled, i.e., $N_\mathcal{S} < K$, as $N_\mathcal{S}$ decreases, the regret due to $R(T)$ decreases until 
$
    1 > \frac{N_\mathcal{S} \OPT}{ K B^\prime},
$
while the term 
$
    \OPT \big( 1- \frac{N_\mathcal{S}}{K} \big)
$
increases. These two terms express two different sources of  regret. More specifically, $R(T)$ decreases in the number of sampled arms because it captures the difference in the performance between our algorithm and the optimal static policy with respect to the portion of the arms that are sampled. The second term, 
$
    \OPT \big( 1 - \frac{N_\mathcal{S}}{K} \big),
$
reflects the loss suffered due to the fact that as the number of sampled arms decreases, the options that are available to our algorithm become fewer. Thus, the algorithm will sometimes miss the opportunity to pick arms with favorable context due to such arms having been discarded. Therefore, the choice of $N_\mathcal{S}$ determines the impact of each of these two sources of regret. Understanding the consequences of this choice can be especially important in applications where for practical reasons operating on the set $[K]$ is infeasible. Despite that our results suggest that the dependence of the regret on the number of arms $K$ may be inevitable, the weakening of this dependence is an open question.

\section{Discussion}\label{sec:discussion}
We have shown that regret sublinear in the number of time periods can be achieved for the problem of linear contextual bandits with knaspacks and clusters of arms. Our approach does not require access to all the available arms, and can thus be utilized in applications where the heterogeneity of the available choices can be meaningfully summarized with clusters, and where individually considering each possible choice is unrealistic, e.g., in online advertising campaigns. It is among our beliefs that the study of forms for summarizing choice heterogeneity in online learning can be a fruitful research direction. Furthermore, our approach can be extended in a number of ways. For instance, the initially sampled arms can serve only as a basis for the clustering, and with additional arms being explored and clustered in later steps of the algorithm. Also, the results in \cite{su2016} suggest that the assumption about knowing the number of the clusters can be relaxed. The improvement of the dependency of the regret on the total number of arms $K$ is a question we find interesting to be explored. Notice that we could have improved this dependency here by allowing the number of sampled arms $N_\mathcal{S}$ to be a function of $K$. However, the budget constraints would then depend on $K$. As far as we are aware, our work does not have negative societal impacts.   

%% file: appendix.tex
The appendix is organized as follows. In Section \ref{sec:app_prel} we state some missing tools from main text. In Section \ref{sec-appendix:clustering} we prove the two lemmas 
about the clustering of the arms. The proofs of the two lemmas presented in Section \ref{subsection:ce} about the confidence ellipsoid can be found in Section \ref{sec-appendix:conf-ellipsoid}. The proof of the lemma in Section \ref{sec:uncertainty} is in Section \ref{sec-appendix:optimism}. In Section \ref{sec-appendix:algorithm} we present the proofs of the results from Section \ref{sec:algorithm} related to the regret of the algorithm.

\section{Preliminary}
\label{sec:app_prel}
In Section~\ref{sec-appendix:prob-tools} we introduce some probability tools. In Section~\ref{subsection:clustering} we introduce the clustering under a mixed linear model. 
\subsection{Probability Tools}
\label{sec-appendix:prob-tools}

\begin{lemma}[Cantelli's inequality]\label{lem:markov}
For random variable $X$ and constant $\xi > 0$,
\begin{equation*}
    \Pr\big[X \geq \E[X] + \xi\big] \leq \frac{Var[X]}{Var[X] + \xi^2}
\end{equation*}
\end{lemma}

\begin{lemma}[Azuma-Hoeffding inequality]
For a martingale $\{ X_i \}_{i = 0, 1, \ldots}$ and constant $\xi > 0$, if $|X_i - X_{i-1}| \leq c_i$ almost surely, then
\begin{equation*}
    \Pr\big[ |X_N - X_0| \geq \xi \big] \leq 2 \exp \Big( - \frac{\xi^2}{2 \sum_{i=1}^N c_i^2} \Big)
\end{equation*}
\end{lemma}

\subsection{Clustering under a Mixed Linear Model}
\label{subsection:clustering}
In order to devise an algorithm that achieves low regret, it will be necessary to  learn the parameters $\bm{\mu}_c$ and $\bm{W}_c$ of each cluster $c \in [C]$. Consequently, we need to assign arms to clusters, with the correctness of an arm's assignment being defined only up to permutation of the cluster identities, as a cluster is defined by its members. However, clustering all of the arms requires obtaining at least $K$ samples, which could be undesirable. Instead, our approach  allows clustering and operating on a subset of the arms. 

While clustering will be performed only for a subset of the $K$ arms, it is important that this subset covers all the $C$ clusters. We initially choose (without playing) at random a subset of the $K$ arms, denoted by $\mathcal{S} \subseteq [K]$, with $|\mathcal{S}| = N_{\mathcal{S}}$. The set of arms $[K] \setminus \mathcal{S}$ that is not chosen will be discarded for all the $T$ time periods. Of course, the regret will be later derived given that the benchmark static policy has all the $K$ arms in its availability. The following result indicates the size of the set $\mathcal{S}$ needed in order to cover all of the clusters with a given probability.

\begin{lemma}
For parameter $\delta > 0$, if the set $\mathcal{S}$ is formed by sampling
$
    N_{\mathcal{S}} = O \big( p^{-1}_{\min} ( T^\delta + \log C) \big)
$
 arms, where each arm is sampled with equal probability, then $\mathcal{S}$ covers the clusters, i.e., $\cup_{a \in S} \{ c(a) \} = [C]$, with probability at least $1 -  O(T^{-2\delta})$.
\label{lemma:unequal-prob-coupon}
\end{lemma} 

The proof of the above lemma relies on arguments similar to those that can be used for the coupon collector problem \cite{flajolet1992}. Once the set $\mathcal{S}$ is determined, in order to collect observations to perform clustering, each arm in $\mathcal{S}$ is played $T_0$ times, with the precise value of $T_0$ being defined later in this subsection. Even though a consumption vector is also observed every time an arm is played, for the purposes of the clustering only, we will utilize just the context and the reward, but we could as well have chosen to utilize one of the $d$ resources instead of the reward. In any case, the resource consumption due to these initial $N_{\mathcal{S}} \cdot T_0$ plays will still have to be subtracted from the budget. 

The arms in $\mathcal{S}$ are clustered using the classifier-Lasso method proposed by \cite{su2016}, which relies on the following objective function
\begin{align}\label{eq:Q}
    & ~ Q\big( (\bm{\mu}_a)_{a \in \mathcal{S}}, (\bm{\mu}_c)_{c \in [C]} \big) \notag \\
    = & ~ \frac{1}{N_{\mathcal{S}} \cdot T_0} \sum_{a \in \mathcal{S}} \sum_{t : a_t = a} \frac{1}{2} \big( r_t(a) - \bm{\mu}_a^\top \bm{x}_{t}(a) \big)^2 \notag\\
     + & ~ \frac{\lambda_1}{N_{\mathcal{S}}} \sum_{a \in \mathcal{S}} \prod_{c \in [C]} \| \bm{\mu}_a - \bm{\mu}_c \|
\end{align}
\noindent where $\lambda_1 \in \mathbb{R}_+$ is a regularization parameter. It is valuable to point out that $N_{\mathcal{S}} + C$ vectors of parameters are estimated in total, since the method does not impose the parameter of an arm to be corresponding to one of the cluster parameters. The estimation of the reward parameters follows by minimizing this objective function,
\begin{align}\label{eq:criterion-clustering}
 & ~\big( (\bm{\hat{\mu}}_a)_{a \in \mathcal{S}}, (\bm{\hat{\mu}}_c)_{c \in [C]} \big) \notag \\
    := &  ~ \argmin_{\big( (\bm{\mu}_a)_{a \in \mathcal{S}}, (\bm{\mu}_c)_{c \in [C]} \big)} Q\big( (\bm{\mu}_a)_{a \in \mathcal{S}}, (\bm{\mu}_c)_{c \in [C]} \big).
\end{align}
 Then, the arm $a \in S$ is clustered as
\begin{align}
    \hat{c}(a) := \sum_{c \in [C]} c \cdot \mathbbm{1}[\bm{\hat{\mu}_a} = \bm{\hat{\mu}_c}]
    \label{eq:clustering-method}
\end{align}
\noindent Since under the classifier-Lasso method it is possible for an arm's estimated parameter to not be equal to the estimated parameter of any cluster, an arm can be assigned to none of the clusters, leading to the case where $\hat{c}(a) = 0$. However, as shown in the proof of the focal result of this subsection, only few of the arms are not assigned to any cluster. 

Besides the coverage of the clusters, we also want to achieve high clustering accuracy. We define the clustering error for cluster $c \in [C]$ as
\begin{align}
    \epsilon_c := \frac{\sum_{a \in \mathcal{S}} \mathbbm{1}[\hat{c}(a) = c , c(a) \neq c]}{\sum_{a \in \mathcal{S}} \mathbbm{1}[\hat{c}(a) = c]}
    \label{eq:clustering-error}
\end{align}
\noindent Thus, the clustering error for $c$ is the proportion of arms assigned to $c$ that should have been assigned to other clusters. We want to ensure that for each cluster $c$, the output of Eq. (\ref{eq:clustering-method}) results to low error $\epsilon_c$. Towards this end, we  make the following assumptions.

\begin{assumption}[Separation]
$\| \bm{\mu}_c - \bm{\mu}_{c^\prime} \| \geq \xi_1 > 0$, for $c, c^\prime \in [C], c \neq c^\prime$.
\label{assumption:separation}
\end{assumption}

\begin{assumption}
The number of clusters $C$ is fixed.
\label{assumption:fixed-C}
\end{assumption}

\begin{assumption}
$T_0 \lambda_1^2 / (\log T_0)^{6 + 2\xi_2} \rightarrow \infty$ and $\lambda_1 (\log T_0)^{\xi_2} \rightarrow 0$, for $\xi_2 > 0$ as $(N_{\mathcal{S}}, T_0) \rightarrow \infty$.
\label{assumption:asymptotics-1}
\end{assumption}

\begin{assumption}
$N_{\mathcal{S}}^{1/2} T_0^{-1} (\log T_0)^9 \rightarrow 0$ and $N_{\mathcal{S}}^2 T_0^{1 - \xi_3 /2} \rightarrow \xi_4 < \infty$, for $\xi_3 \geq 6$ as $(N_{\mathcal{S}}, T_0) \rightarrow \infty$.
\label{assumption:asymptotics-2}
\end{assumption}

Assumption \ref{assumption:separation} is needed to ensure that the clusters are distinguishable, Assumption \ref{assumption:fixed-C} disallows the number of clusters to grow asymptotically, while Assumptions \ref{assumption:asymptotics-1} and \ref{assumption:asymptotics-2} impose appropriate rates on $\lambda_1, N_{\mathcal{S}}$, and $T_0$. Had we chosen to perform clustering based on one of the $d$ resources instead of the reward, then Assumption \ref{assumption:separation} would have been for that resource.

\begin{lemma}
Under Assumptions \ref{assumption:linearity}-\ref{assumption:iid},  \ref{assumption:separation}-\ref{assumption:asymptotics-2}, the clustering error is  $\epsilon_c = o(p_{\min}^{-1} N_{\mathcal{S}}^{-1})$ with high probability, for any $c \in [C]$.
\label{lemma:cluster-error}
\end{lemma}

In the proof of Lemma \ref{lemma:cluster-error} we exploit the fact that $\Pr[\hat{c}(a) = c(a)] = 1 - o(N_\mathcal{S}^{-1})$ (Su et al., 2016, pp. 2250). Since the clustering error is, in asymptotic terms, the same for all clusters, we shall use $\epsilon_c$ to refer to this error for any cluster. Moreover, since $N_\mathcal{S}$ is given by Lemma \ref{lemma:unequal-prob-coupon}, the number of samples required from each arm in $\mathcal{S}$ for the clustering can be specified by satisfying Assumption \ref{assumption:asymptotics-2}.

\begin{claim}
For $T_0 = N_{\mathcal{S}}$, Assumption  \ref{assumption:asymptotics-2} is satisfied.
\end{claim}

\section{Proofs for Section~\ref{sec:app_prel}}
\label{sec-appendix:clustering}
Here in this section, we provide some proofs for Section~\ref{sec:app_prel}. In Section~\ref{sec:prof_unequalprob} we provide the proof of Lemma~\ref{lemma:unequal-prob-coupon}. In Section~\ref{sec:proof_lemma_cluster_error} we provide proof for Lemma~\ref{lemma:cluster-error}. 

\subsection{Proof of Lemma \ref{lemma:unequal-prob-coupon}}
\label{sec:prof_unequalprob}

In this proof, we will treat the portions in $\bm{p}$ as probabilities from a distribution by sampling an arm with replacement, instead of without replacement. Since putting a sampled arm back in the sampling distribution only decreases the probability of the next sample being from a non-sampled cluster, the result we will derive will imply a lower bound for our problem.

Consider the following two-step sampling process, repeated for $j = 1, 2, 3, \ldots$, for the collection of arms: In the first step, a cluster $c \sim \bm{p}$ is drawn and then the arm $a_j$ with $c(a_j) = c$ is considered. In the second step, the arm $a_j$ is kept in the set of collected arms with probability $\frac{p_{\min}}{p_c}$, and it is discarded with probability $1 - \frac{p_{\min}}{p_c}$. Let $l_j$ denote the outcome of the $j^{\text{th}}$ draw, so that $l_j = 0$ if the sampled arm was discarded and $l_j = c(a_j)$ if the arm was kept.  Under this two-step sampling process, we have for $c \in [C]$ that
\begin{equation*}
    \Pr[l_j = c] = p_c \frac{p_{\min}}{p_c} = p_{\min}
\end{equation*}

\noindent and so, the corresponding sampling distribution for a draw is
\begin{align*}
    \bm{p}_0 &:= \big( \Pr[l_j = 0], \Pr[l_j = 1], \ldots, \Pr[l_j = C] \big)
    \\
    &= \big( 1 - C p_{\min}, p_{\min}, \ldots, p_{\min} \big) 
\end{align*}

Thus, with regard to $[C]$, $\bm{p}_0$ is a uniform distribution. Then, the probability of the event that a new draw gives a new cluster when $i-1$ clusters have already been collected is $\frac{C-i+1}{C}(1 - (1-C p_{\min})) = (C-i+1) p_{\min}$. Let $L$ denote the number of draws needed to cover all of the clusters. From properties of the geometric distribution it follows that
\begin{align*}
     \E[L] &=  p_{\min}^{-1} \sum_{c \in [C]} \frac{1}{C - c + 1}
     \\
     &= O( p_{\min}^{-1} \log C )
\end{align*}
and
\begin{align*}
    Var[L] &< p_{\min}^{-2} \sum_{c \in [C]} \frac{1}{(C - c + 1)^2}
    \\
    &< 2 p_{\min}^{-2}
\end{align*}

By Cantelli's inequality we get
\begin{align*}
    \Pr[L \leq \E[L] + p^{-1}_{\min} T^\delta] &\geq 1 - \frac{Var[L]}{Var[L] + p^{-2}_{\min} T^{2\delta}}
    \\
    &> 1 - \frac{1}{1 + \frac{1}{2}T^{2\delta}}
\end{align*}

\subsection{Proof of Lemma \ref{lemma:cluster-error}}
\label{sec:proof_lemma_cluster_error}

In this proof, we will exploit the Azuma-Hoeffding inequality and a result from \cite{su2016} that holds under the stated assumptions.

\begin{lemma}[\cite{su2016}, pp. 2250] $\Pr[\hat{c}(a) = c(a)] = 1 - o(N_{\mathcal{S}}^{-1})$
\label{lemma:su2016-1}
\end{lemma}

Now, for the clustering error $\epsilon_c$ we have
\begingroup
\allowdisplaybreaks
\begin{align}
    \epsilon_c &= \frac{\sum_{a \in \mathcal{S}} \mathbbm{1}[\hat{c}(a) = c , c(a) \neq c]}{\sum_{a \in \mathcal{S}} \mathbbm{1}[\hat{c}(a) = c]}
    \notag \\
    &=  \frac{\sum_{a \in \mathcal{S}} \mathbbm{1}[\hat{c}(a) = c, c(a) \neq c]}{\sum_{a \in \mathcal{S}} \mathbbm{1}[c(a) = c] - \sum_{a \in \mathcal{S}} \mathbbm{1}[\hat{c}(a) \neq c, c(a) = c] + \sum_{a \in \mathcal{S}} \mathbbm{1}[\hat{c}(a) = c, c(a) \neq c] }
    \notag \\
    &\leq \frac{\sum_{a \in \mathcal{S}} \mathbbm{1}[\hat{c}(a) = c, c(a) \neq c]}{\sum_{a \in \mathcal{S}} \mathbbm{1}[c(a) = c] - \sum_{a \in \mathcal{S}} \mathbbm{1}[\hat{c}(a) \neq c, c(a) = c] }
    \notag \\
    &= \frac{\sum_{a \in \mathcal{S}} \mathbbm{1}[\hat{c}(a) = c, c(a) \neq c]}{\sum_{a: c(a) = c} 1 - \mathbbm{1}[\hat{c}(a) \neq c]}
    \notag \\
    &\leq \frac{\sum_{a \in \mathcal{S}} \mathbbm{1}[\hat{c}(a) = c, c(a) \neq c]}{O(1) \E\big[\sum_{a: c(a) = c} 1 - \mathbbm{1}[\hat{c}(a) \neq c]\big]}
    \label{eq:1-26-A}
    \\
    &= \frac{\sum_{a \in \mathcal{S}} \mathbbm{1}[\hat{c}(a) = c , c(a) \neq c]}{O(1) \E\big[\sum_{a \in \mathcal{S}} \mathbbm{1}[c(a) = c, \hat{c}(a) = c]\big]}
    \notag \\
    &= \frac{\sum_{a \in \mathcal{S}} \mathbbm{1}[\hat{c}(a) = c, c(a) \neq c]}{O(1) \sum_{a \in \mathcal{S}} \Pr[c(a) = c] \Pr[\hat{c}(a) = c | c(a) = c]  }
    \notag \\
    &= \frac{\sum_{a \in \mathcal{S}} \mathbbm{1}[\hat{c}(a) = c, c(a) \neq c]}{O(1) N_{\mathcal{S}} p_c (1 - o(N_{\mathcal{S}}^{-1}) ) }
   \notag  \\
    &\leq \frac{\sum_{a \in \mathcal{S}} \mathbbm{1}[\hat{c}(a) = c, c(a) \neq c]}{O(1) N_{\mathcal{S}} p_{\min} (1 - o(N_{\mathcal{S}}^{-1}) ) }
    \notag \\
    &\leq O(1) \frac{\E\big[\sum_{a \in \mathcal{S}} \mathbbm{1}[\hat{c}(a) = c, c(a) \neq c]\big]}{N_{\mathcal{S}} p_{\min} (1 - o(N_{\mathcal{S}}^{-1}) )}
    \label{eq:1-26-B}
    \\
    &= O(1) \frac{\sum_{a \in \mathcal{S}} \Pr[c(a) \neq c] \Pr[\hat{c}(a) = c | c(a) \neq c] }{N_{\mathcal{S}} p_{\min} (1 - o(N_{\mathcal{S}}^{-1}) )}
    \notag \\
    &\leq O(1) \frac{\sum_{a \in \mathcal{S}} \Pr[c(a) \neq c] \Pr[\hat{c}(a) \neq c(a)] }{N_{\mathcal{S}} p_{\min} (1 - o(N_{\mathcal{S}}^{-1}) )}
    \notag \\
    &= O(1) \frac{N_{\mathcal{S}} (1- p_c) o(N_{\mathcal{S}}^{-1})}{N_{\mathcal{S}} p_{\min} (1 - o(N_{\mathcal{S}}^{-1}) )}
    \notag \\
    &= O(1) \frac{(1- p_c) o(N_{\mathcal{S}}^{-1})}{p_{\min} (1 - o(N_{\mathcal{S}}^{-1}) )}
    \notag \\
    &\leq \frac{o(p_{\min}^{-1} N_{\mathcal{S}}^{-1})}{1 - o(N_{\mathcal{S}}^{-1})}
    \notag \\
    &\leq o(p_{\min}^{-1} N_{\mathcal{S}}^{-1}) \notag 
\end{align}
\endgroup
where Eq. (\ref{eq:1-26-A}) and (\ref{eq:1-26-B}) follow from the  Azuma-Hoeffding inequality.

\section{Proofs for Section~\ref{sec:preliminary}}
\label{sec-appendix:conf-ellipsoid}
Here in this section, we provide some missing proofs for Section~\ref{sec:preliminary}. In Section~\ref{sec:proof_ellipsoid} we provide proof for Lemma~\ref{lemma:ellipsoid}. In Section~\ref{sec:proof_sum_reward} we provide proof for Lemma~\ref{lemma:sum-rewards}. 
\subsection{Proof of Lemma  \ref{lemma:ellipsoid}}
\label{sec:proof_ellipsoid}

For notational convenience, let
\begin{definition}
 \begin{align*}
     \bm{X} &:= (\bm{x}_1^\top, \ldots, \bm{x}_t^\top) \in [0, 1]^{t \times m}
     \\
     \bm{y} &:= (y_1, \ldots, y_t) \in [0, 1]^t
     \\
     \bm{\eta} &:= (\eta_1, \ldots, \eta_t) \in \mathbb{R}^t
     \\
     \bm{s}_t &:= \sum_{i=1}^{t-1} \bm{x}_i (\eta_i - \E[\eta_i | \bm{x}_i])
     \\
     \bar{\gamma} &:= \| \bm{\gamma} \|_{\infty} 
     \\
     \langle \bm{a}, \bm{b} \rangle_{\bm{M}} &:= \bm{a}^\top \bm{M} \bm{b}, \text{ for } \bm{a}, \bm{b} \in \mathbb{R}^m, \bm{M} \in \mathbb{R}^{m \times m}
 \end{align*}
\end{definition}

 For the proof we will need the following two Facts and four Lemmas.
 
 \begin{fact}
We can show that $\hat{\bm{\mu}}_t = \bm{\mu} - \lambda_2 \bm{M}_t^{-1} \bm{\mu} +  \bm{M}_t^{-1}  \bm{X}^\top \bm{\eta}$
 \label{fact:10-20-21-A}
\end{fact}
\begin{proof}
\begin{align*}
     \hat{\bm{\mu}}_t &= \bm{M}_t^{-1}  \bm{X}^\top \bm{y}
     \\
     &= \bm{M}_t^{-1}  \bm{X}^\top (\bm{X} \bm{\mu} + \bm{\eta})
     \\
     &= \bm{M}_t^{-1}  \bm{X}^\top \bm{X} \bm{\mu} + \bm{M}_t^{-1}  \bm{X}^\top \bm{\eta}
     \\
     &= \bm{M}_t^{-1}  \bm{X}^\top \bm{X} \bm{\mu} + \bm{M}_t^{-1} \lambda_2 \bm{I}_m \bm{\mu} - \bm{M}_t^{-1} \lambda_2 \bm{I}_m \bm{\mu}  + \bm{M}_t^{-1}  \bm{X}^\top \bm{\eta}
     \\
     &=  \bm{M}_t^{-1}  \bm{M}_t \bm{\mu} - \lambda_2  \bm{M}_t^{-1} \bm{\mu} +  \bm{M}_t^{-1}  \bm{X}^\top \bm{\eta}
     \\
     &=\bm{\mu} - \lambda_2 \bm{M}_t^{-1} \bm{\mu} +  \bm{M}_t^{-1}  \bm{X}^\top \bm{\eta}
 \end{align*}
 Thus, we complete the proof.
\end{proof}

\begin{fact}
For any $\bm{x} \in \mathbb{R}^m$, we have that $\bm{x}^\top \hat{\bm{\mu}}_t -  \bm{x}^\top \bm{\mu} = \langle  \bm{x}, \bm{X}^\top \bm{\eta} \rangle_{\bm{M}_t^{-1}} - \lambda_2 \langle \bm{x}, \bm{\mu} \rangle_{\bm{M}_t^{-1}}$
\label{fact:10-20-21-B}
\end{fact}
\begin{proof}
From Fact \ref{fact:10-20-21-A} we have
  \begin{align*}
      \bm{x}^\top \hat{\bm{\mu}}_t -  \bm{x}^\top \bm{\mu} &= \bm{x}^\top \big( \bm{\mu} - \lambda_2  \bm{M}_t^{-1}  \bm{\mu}  +  \bm{M}_t^{-1}  \bm{X}^\top \bm{\eta}  \big) -  \bm{x}^\top \bm{\mu}
      \\
      &= \bm{x}^\top \bm{M}_t^{-1}  \bm{X}^\top \bm{\eta} - \lambda_2  \bm{x}^\top  \bm{M}_t^{-1}  \bm{\mu}
      \\
      &= \langle  \bm{x}, \bm{X}^\top \bm{\eta} \rangle_{\bm{M}_t^{-1}} - \lambda_2 \langle \bm{x}, \bm{\mu} \rangle_{\bm{M}_t^{-1}}
  \end{align*}
  Thus, we complete the proof.
\end{proof}
 
 \begin{lemma}
We can show that
\begin{align*}
    |\bm{x}^\top \hat{\bm{\mu}}_t -  \bm{x}^\top \bm{\mu}|  \leq \| \bm{x} \|_{\bm{M}_t^{-1}} \Big( \| \bm{s_t} \|_{\bm{M}_t^{-1}} +  \epsilon \|  \bm{X}^\top \bm{X} \bm{\gamma} \|_{\bm{M}_t^{-1}} + \sqrt{\lambda_2 m} \ \Big)
\end{align*}
\label{lemma:10-20-21-A}
\end{lemma}
\begin{proof}
Using the Cauchy-Schwarz inequality on Fact \ref{fact:10-20-21-B}, we have
\begin{align*}
    |\bm{x}^\top \hat{\bm{\mu}}_t -  \bm{x}^\top \bm{\mu}| &\leq \| \bm{x} \|_{\bm{M}_t^{-1}} \Big( \| \bm{X}^\top \bm{\eta} \|_{\bm{M}_t^{-1}} + \lambda_2 \| \bm{\mu} \|_{\bm{M}_t^{-1}} \Big)
\end{align*}
We can upper bound the second term of the above equation as follows:
\begin{align*}
       \| \bm{X}^\top \bm{\eta} \|_{M_t^{-1}} + \lambda_2 \| \bm{\mu} \|_{M_t^{-1}} 
    & \leq    \| \bm{X}^\top \bm{\eta} \|_{\bm{M}_t^{-1}} + \sqrt{\lambda}_2 \| \bm{\mu} \|_2
    \label{eq:9-23-21-A}
    \\
    & \leq   \| \bm{X}^\top \bm{\eta} \|_{\bm{M}_t^{-1}} + \sqrt{\lambda_2 m} 
    \\
    &=   \| \bm{X}^\top (\bm{\eta} + \E[\bm{\eta}|\bm{X}] - \E[\bm{\eta}|\bm{X}] ) \|_{\bm{M}_t^{-1}} + \sqrt{\lambda_2 m} 
    \\
    & =   \| \bm{s_t} + \bm{X}^\top  \E[\bm{\eta}|\bm{X}] \|_{\bm{M}_t^{-1}} + \sqrt{\lambda_2 m} \ 
    \\
    & =    \| \bm{s_t} + \epsilon \bm{X}^\top \bm{X} \bm{\gamma} \|_{\bm{M}_t^{-1}} + \sqrt{\lambda_2 m} \ 
    \\
    & \leq   \| \bm{s_t} \|_{\bm{M}_t^{-1}} +  \|\epsilon  \bm{X}^\top \bm{X} \bm{\gamma} \|_{\bm{M}_t^{-1}} + \sqrt{\lambda_2 m} \  
    \\
    & =   \| \bm{s_t} \|_{\bm{M}_t^{-1}} +  \epsilon \|  \bm{X}^\top \bm{X} \bm{\gamma} \|_{\bm{M}_t^{-1}} + \sqrt{\lambda_2 m} \  
\end{align*}

where the second inequality follows from that $\| \bm{\mu} \|_{\bm{M}_t^{-1}}^2 \leq \frac{1}{\lambda_{\min} (\bm{M}_{t}) \| \bm{\mu} \|_2^2 } \leq \frac{1}{\lambda_2  \| \bm{\mu} \|_2^2 }$. 
\end{proof}

\begin{lemma}\label{lemma:sub-gaussian}
Consider the $\sigma-$algebra $F_t = \sigma(\bm{x}_1, \ldots, \bm{x}_t, \eta_1, \ldots, \eta_{t-1})$, such that  $\{ F_t \}_{t=1}^\infty$ is a filtration, and $\eta_t - \E[\eta_t |\bm{x}_t]$ is $F_t$-measurable. Then, $\eta_t - \E[\eta_t |\bm{x}_t] $  is $(R+1)$-sub-Gaussian.
\end{lemma}
\begin{proof}
It suffices to show that $\eta_t - \E[\eta_t | \bm{x}_t]$ lies in an interval of length at most $2(R+1)$. We can upper bound $|\eta_t - \E[\eta_t | \bm{x}_t]|$ as follows:
\begin{align*}
    |\eta_t - \E[\eta_t | \bm{x}_t]| &= |u_t + h_t - \epsilon\bm{\gamma}^\top \bm{x}_t|
    \\
    &\leq |u_t| + |h_t| + |\epsilon \bm{\gamma}^\top \bm{x}_t|
    \\
    &\leq 2R + (1+\epsilon) |\bm{\gamma}^\top \bm{x}_t|
    \\
    &\leq 2(R+1)
\end{align*}
\end{proof}

 \begin{lemma}\label{lemma:9-30-known-AY-1}
 For any $\zeta \in (0, 1)$, with probability at least $1 - \zeta$, if $\eta_t - \E[\eta_t | \bm{x}_t]$ is $(R+1)$-sub-Gaussian, then for all $t > 0$,
 \begin{equation*}
     \| \bm{s}_t \|^2_{\bm{M}_t^{-1}} \leq 2 (R+1)^2 \log \big(  \det(\bm{M}_t)^{1/2} \det(\lambda_2 \bm{I}_m)^{-1/2}/\zeta \big)
 \end{equation*}
\end{lemma}

\begin{proof}
We apply Theorem 1 of \cite{abbasi2011} with $\eta_t - \E[\eta_t | \bm{x}_t]$ in the place of $\eta_t$.
\end{proof}

\begin{lemma}\label{lemma:10-20-B}
We can show that
\begin{align*}
    \| \bm{X}^\top \bm{X} \bm{\gamma} \|_{\bm{M}_t^{-1}} \leq \bar{\gamma}  m  \sqrt{t}
\end{align*}
\end{lemma}

\begin{proof}
Since $\bm{X}^\top \bm{X} \in \R^{m \times m}$ contains non-negative entries, we have
\begin{align*}
    \| \bm{X}^\top \bm{X} \bm{\gamma} \|_{\bm{M}_t^{-1}} 
    \leq \bar{\gamma} \cdot \| \bm{X}^\top \bm{X} \bm{1}_m \|_{\bm{M}_t^{-1}}
\end{align*}
Next, we just need to upper bound $\| \bm{X}^\top \bm{X} \bm{1}_m \|_{\bm{M}_t^{-1}}^2$. By the properties of PSD/PD matrices, we know that
\begin{align*}
    \lambda_2 \bm{I}_m + \bm{X}^\top \bm{X} \succ  \bm{X}^\top \bm{X} \succeq 0
\end{align*} 
which implies that 
\begin{align*}
  (\bm{X}^\top \bm{X})^{1/2} \cdot (\lambda_2 \bm{I}_m + \bm{X}^\top \bm{X})^{-1} \cdot (\bm{X}^\top \bm{X})^{1/2} \prec \bm{I}_m
\end{align*} 

Thus, we have
\begin{align*}
     \bm{1}_m^\top \bm{X}^\top \bm{X}  (\lambda_2 \bm{I}_m + \bm{X}^\top \bm{X})^{-1} \bm{X}^\top \bm{X} \bm{1}_m 
    \leq & ~  \bm{1}_m^\top \bm{X}^\top \bm{X}  \bm{1}_m 
    \\
    = & ~ \| \bm{X}  \bm{1}_m  \|_2^2 \\
    \leq & ~  t m^2
\end{align*}
where the last step follows since each entry of $\bm{X}$ is between $0$ and $1$.
\end{proof}

 The rest of the proof follows that of Theorem 2 in \cite{abbasi2011}. By lemmas \ref{lemma:10-20-21-A},   \ref{lemma:sub-gaussian}, \ref{lemma:9-30-known-AY-1}, and \ref{lemma:10-20-B},   we have that for any $\zeta \in (0, 1)$, with probability at least $1 - \zeta$,

\begin{equation*}
    |\bm{x}^\top \hat{\bm{\mu}}_t -  \bm{x}^\top \bm{\mu}| \leq \| \bm{x} \|_{\bm{M}_t^{-1}} \cdot A_1  
\end{equation*}
where
\begin{align*}
    A_1: = \Big((R+1) \sqrt{ 2 \log \big(  \det(\bm{M}_t)^{1/2} \det(\lambda_2 \bm{I}_m)^{-1/2}/\zeta \big) } + \epsilon \bar{\gamma} m \sqrt{t}  + \sqrt{\lambda_2 m} \Big) 
\end{align*}

Now, by letting $\bm{x} = \bm{M}_t (\hat{\bm{\mu}}_t - \bm{\mu})$, we have

\begin{align*}
    \| \hat{\bm{\mu}}_t - \bm{\mu} \|^2_{\bm{M}_t} &\leq \| \bm{M}_t (\hat{\bm{\mu}}_t - \bm{\mu}) \|_{\bm{M}_t^{-1}} \cdot A_1
    \\
    &= \| \hat{\bm{\mu}}_t - \bm{\mu} \|_{\bm{M}_t} \cdot A_1
\end{align*}

Dividing both sides by $\| \hat{\bm{\mu}}_t - \bm{\mu} \|_{\bm{M}_t}$ we get 

\begin{align}
    \| \hat{\bm{\mu}}_t - \bm{\mu} \|_{\bm{M}_t} &\leq A_1
\end{align}
 
 Since $\| \bm{x}_t \|_2 \leq \sqrt{m}$, and $\bm{M}_t$ and $\lambda_2 \bm{I}_m$ are positive-definite matrices, we can upper bound the first term in $A_1$ (ignoring the term $R+1$) as follows
 
 \begin{equation}
     \sqrt{ 2 \log \big(  \det(\bm{M}_t)^{1/2} \det(\lambda_2 \bm{I}_m)^{-1/2}/\zeta \big) } \leq \sqrt{ m \log \Big( \frac{1 + t m /\lambda_2}{\zeta} \Big) }
 \end{equation}
 
 and thus
 
 \begin{align}
    \| \hat{\bm{\mu}}_t - \bm{\mu} \|_{\bm{M}_t} &\leq (R+1) \sqrt{ m \log \Big( \frac{1 + t m /\lambda_2}{\zeta} \Big) } + \epsilon \bar{\gamma} m \sqrt{t} + \sqrt{\lambda_2 m} 
\end{align}

where $\bar{\gamma} \leq 1$.

\subsection{Proof of Lemma \ref{lemma:sum-rewards}}
\label{sec:proof_sum_reward}

We will need the following fact and lemma.

\begin{fact} For any positive definite matrix $\bm{M} \in \mathbb{R}^{m \times m}$ and any two vectors $\bm{a}, \bm{b} \in \mathbb{R}^m$, it holds that $|\bm{a}^\top \bm{b}| \leq \|\bm{a}\|_{\bm{M}} \| \bm{b} \|_{\bm{M}^{-1}}$.
\label{fact:9-30-known-7}
\end{fact}

 \begin{lemma}[Lemma 3 of \cite{agrawal2016a}]
For $\bm{x}_i \in \mathbb{R}^m$ with $\| \bm{x}_i \|_2 \leq \sqrt{m}$, it holds that
\begin{equation*}
     \sum_{i=1}^t \| \bm{x}_i \|_{\bm{M}_i^{-1}} \leq \sqrt{m t \log(t)}
\end{equation*}
\label{lemma:norm-known-3}
\end{lemma}

Now, we derive the statement of the lemma as:
\begin{align*}
    \sum_{t=1}^T | \bm{\tilde{\mu}}_t^\top \bm{x}_t - \bm{\mu}^\top \bm{x}_t | &\leq \sum_{t=1}^T  \| \bm{x}_t \|_{\bm{M}_t^{-1}} \| \bm{\tilde{\mu}}_t - \bm{\mu} \|_{\bm{M}_t}
    \\
    & \leq \Big( 3 \sqrt{ m \log \big(  T m/ \zeta \big) } + \epsilon m \sqrt{T} + \sqrt{m} \Big) \sum_{t=1}^T  \| \bm{x}_t \|_{\bm{M}_t^{-1}}
    \\
    & \leq \Big( 4 \sqrt{ m \log \big(  T m/ \zeta \big) } + \epsilon  m \sqrt{T} \Big) \sum_{t=1}^T  \| \bm{x}_t \|_{\bm{M}_t^{-1}}
    \\
    & \leq \Big( 4 \sqrt{ m \log \big(  T m/ \zeta \big) } + \epsilon m \sqrt{T} \Big) \sqrt{m T \log(T)}
    \\
    & = 4 m \sqrt{ T \log \big(  T m/ \zeta \big) \log(T)  } + \epsilon  m^{\frac{3}{2}} T \sqrt{\log(T)} 
\end{align*}
where the first step follows from Fact \ref{fact:9-30-known-7}, the second from Lemma \ref{lemma:ellipsoid}, and the fourth from Lemma \ref{lemma:norm-known-3}.

\section{Proofs for Section~\ref{sec:uncertainty}}
\label{sec-appendix:optimism}
Here we provide the following proof. 
\subsection{Proof of Lemma \ref{lemma:optimism}}

Statements $a)$ and $b)$ follow directly from Eq. (\ref{eq:optimistic-mu}) and (\ref{eq:optimistic-W}). For statement $c)$ we have: 

\begin{align*}
    |\sum_{i=N_{\mathcal{S}} \cdot T_0 + 1}^t  \bm{x}_i(a_i)^\top ( \tilde{\bm{\mu}}_{a_i, i}  -  {\bm{\mu}}_{\hat{c}(a_i)} ) |
    &\leq
    \sum_{i=N_{\mathcal{S}} \cdot T_0 + 1}^t |  \bm{x}_i(a_i)^\top (\tilde{\bm{\mu}}_{a_i, i}  -  {\bm{\mu}}_{\hat{c}(a_i)} ) | 
    \\
    &= \sum_{c \in [C]} \sum_{\substack{i: \hat{c}(a_i) = c \\ i > N_{\mathcal{S}} \cdot T_0}} | \bm{x}_i(a_i)^\top (  \tilde{\bm{\mu}}_{a_i, i} -  \bm{\mu}_c) |
    \\
    &\leq \sum_{c \in [C]} 4m \sqrt{t_c \log \big(  t_c m / \zeta \big) \log(t_c)} + \epsilon_c  m^\frac{3}{2} t_c \sqrt{\log(t_c)}
    \\
    &\leq 4 C m \sqrt{t \log \big( t m / \zeta \big) \log(t)} + \epsilon_c  m^\frac{3}{2} t \sqrt{\log(t)}
\end{align*}
where the first step follows from the triangle inequality, and the third from Lemma \ref{lemma:sum-rewards}. Statement $d)$ follows similarly.

\section{Proofs for Section~\ref{sec:algorithm}}
\label{sec-appendix:algorithm}

Here we provide the following proof. In Section~\ref{sec:algorithm_appendix:proof_lemma_Z} we provide the proof for Lemma~\ref{lemma:Z}. In Section~\ref{sec:algorithm_appendix:proof_Theorem_main} we provide the proof for Theorem~\ref{theorem:main}. 

\subsection{Proof of Lemma \ref{lemma:Z}}\label{sec:algorithm_appendix:proof_lemma_Z}

For notational convenience, let
\begin{equation*}
    \rho(\pi) := \frac{K}{N_\mathcal{S}^2 T_0}  \sum_{t=1}^{N_{\mathcal{S}}  T_0} \sum_{a \in \mathcal{S}} \hat{\bm{\mu}}_{\hat{c}(a), N_{\mathcal{S}}  T_0+1}^\top \bm{x}_t(a) \pi(a, \bm{X}_t)
\end{equation*}

Denote the maximizer of the program of the lemma by $\pi'$, i.e.,
\begin{align*}
    \pi' &:= \argmax_\pi \rho(\pi)
    \\
    \text{ s.t. } & \frac{K}{N_\mathcal{S}^2 T_0} \sum_{t=1}^{N_{\mathcal{S}}  T_0}  \sum_{a \in \mathcal{S}} \hat{\bm{W}}_{\hat{c}(a), N_{\mathcal{S}}  T_0+1}^\top \bm{x}_t(a) \pi(a, \bm{X}_t)  \leq \frac{B}{T} \bm{1}_d
\end{align*}

By the definition of the benchmark policy in Section \ref{sec:problem}, we have $\pi^* =  \argmax_{\pi \in \Pi} r(\pi) \text{ subject to } \bm{v}(\pi) \leq \frac{B}{T} \cdot \bm{1}_d$. For ease of exposition, we will first ignore the constraints and account for them later. Notice that we want to prove that the difference $|r(\pi^*) - \rho(\pi')|$ is small. The next lemma allows us to work with a more convenient term instead. 

\begin{lemma}
If $\max_{\pi \in \{ \pi^*, \pi' \}} | r(\pi) - \rho(\pi) | \leq \epsilon'$ then $| r(\pi^*) - \rho(\pi')| \leq \epsilon'$.
\end{lemma}
\begin{proof}
Considering $\pi^*$ we have
\begin{equation*}
    r(\pi^*) - \rho(\pi^*) \leq \epsilon'
\end{equation*}
which implies that
\begin{equation*}
    r(\pi^*) - \rho(\pi') \leq \epsilon'
\end{equation*}
and by considering $\pi'$ we similarly get
\begin{equation*}
    \rho(\pi') - r(\pi^*) \leq \epsilon'
\end{equation*}
Thus, it follows that $| r(\pi^*) - \rho(\pi')| \leq \epsilon'$.
\end{proof}

Therefore, it suffices to prove that the difference $\max_{\pi \in \{ \pi^*, \pi' \}} | r(\pi) - \rho(\pi) |$ is small. We first consider $\pi^*$ and then $\pi'$.

\begin{lemma}\label{lemma:1-27-21}
$|r(\pi^*) - \rho(\pi^*)| < o(1) $
\end{lemma}
\begin{proof}
\begin{align*}
    & |r(\pi^*) - \rho(\pi^*)| \\
    &\leq | \E_{\bm{X}}\Big[  \sum_{a \in [K]} \bm{\mu}_{c(a)}^\top \bm{x}(a) \pi^*(a, \bm{X}) \Big] -  \frac{K}{N_\mathcal{S}} \E_{\bm{X}}\Big[  \sum_{a \in \mathcal{S}} \bm{\mu}_{c(a)}^\top \bm{x}(a) \pi^*(a, \bm{X}) \Big] |
    \\
    &+ | \frac{K}{N_\mathcal{S}} \E_{\bm{X}}\Big[  \sum_{a \in \mathcal{S}} \bm{\mu}_{c(a)}^\top \bm{x}(a) \pi^*(a, \bm{X}) \Big] - \frac{K}{N_\mathcal{S}} \E_{\bm{X}}\Big[  \sum_{a \in \mathcal{S}} \bm{\mu}_{\hat{c}(a)}^\top \bm{x}(a) \pi^*(a, \bm{X}) \Big] |
    \\
    &+ |\frac{K}{N_\mathcal{S}} \E_{\bm{X}}\Big[  \sum_{a \in \mathcal{S}} \bm{\mu}_{\hat{c}(a)}^\top \bm{x}(a) \pi^*(a, \bm{X}) \Big] - \frac{K}{N_\mathcal{S}} \E_{\bm{X}}\Big[  \sum_{a \in \mathcal{S}} \hat{\bm{\mu}}_{\hat{c}(a), N_\mathcal{S}T_0+1}^\top \bm{x}(a) \pi^*(a, \bm{X}) \Big] |
    \\
    &+ | \frac{K}{N_\mathcal{S}} \E_{\bm{X}}\Big[  \sum_{a \in \mathcal{S}} \hat{\bm{\mu}}_{\hat{c}(a), N_\mathcal{S}T_0+1}^\top \bm{x}(a) \pi^*(a, \bm{X}) \Big] - \frac{K}{N_\mathcal{S}} \frac{1}{N_\mathcal{S} T_0} \sum_{t=1}^{N_\mathcal{S} T_0} \sum_{a \in \mathcal{S}} \hat{\bm{\mu}}_{\hat{c}(a), N_\mathcal{S}T_0+1}^\top \bm{x}(a) \pi^*(a, \bm{X}) \Big] |
    \\
    &< o(1) 
\end{align*}
The first difference is zero, as the set $\mathcal{S}$ is sampled uniformly at random. The second difference is $o(1)$ because the probability of clustering an arm incorrectly is $o(N^{-1}_\mathcal{S})$. The third difference is $o(1)$ because the regularized ordinary least squares estimator is consistent when there is no clustering error, i.e., $\hat{\bm{\mu}}_{c, N_\mathcal{S} T_0 + 1} \xrightarrow{p} \bm{\mu}_c$ as $(N_\mathcal{S}, T_0) \xrightarrow{} \infty$ (and as long as the set $[C]$ is covered by $\mathcal{S}$), and the clustering error vanishes asymptotically. The fourth difference is $o(1)$ because it is the difference between an expectation and its empirical counterpart.
\end{proof}

\begin{lemma}
$|r(\pi') - \rho(\pi')| < o(1) $
\end{lemma}
\begin{proof}
\begin{align*}
    & ~ |r(\pi') - \rho(\pi')| \\
    &\leq | \E_{\bm{X}}\Big[  \sum_{a \in [K]} \bm{\mu}_{c(a)}^\top \bm{x}(a) \pi'(a, \bm{X}) \Big] -  \frac{K}{N_\mathcal{S}} \E_{\bm{X}}\Big[  \sum_{a \in \mathcal{S}} \bm{\mu}_{c(a)}^\top \bm{x}(a) \pi'(a, \bm{X}) \Big] |
    \\
    &+ | \frac{K}{N_\mathcal{S}} \E_{\bm{X}}\Big[  \sum_{a \in \mathcal{S}} \bm{\mu}_{c(a)}^\top \bm{x}(a) \pi'(a, \bm{X}) \Big] - \frac{K}{N_\mathcal{S}} \E_{\bm{X}}\Big[  \sum_{a \in \mathcal{S}} \bm{\mu}_{\hat{c}(a)}^\top \bm{x}(a) \pi'(a, \bm{X}) \Big] |
    \\
    &+ |\frac{K}{N_\mathcal{S}} \E_{\bm{X}}\Big[  \sum_{a \in \mathcal{S}} \bm{\mu}_{\hat{c}(a)}^\top \bm{x}(a) \pi'(a, \bm{X}) \Big] - \frac{K}{N_\mathcal{S}} \E_{\bm{X}}\Big[  \sum_{a \in \mathcal{S}} \hat{\bm{\mu}}_{\hat{c}(a), N_\mathcal{S}T_0+1}^\top \bm{x}(a) \pi'(a, \bm{X}) \Big] |
    \\
    &+ | \frac{K}{N_\mathcal{S}} \E_{\bm{X}}\Big[  \sum_{a \in \mathcal{S}} \hat{\bm{\mu}}_{\hat{c}(a), N_\mathcal{S}T_0+1}^\top \bm{x}(a) \pi'(a, \bm{X}) \Big] - \frac{K}{N_\mathcal{S}} \frac{1}{N_\mathcal{S} T_0} \sum_{t=1}^{N_\mathcal{S} T_0} \sum_{a \in \mathcal{S}} \hat{\bm{\mu}}_{\hat{c}(a), N_\mathcal{S}T_0+1}^\top \bm{x}(a) \pi'(a, \bm{X}) \Big] |
    \\
    &< o(1)
\end{align*}
Each difference is $o(1)$ following arguments similar to those in the proof of Lemma \ref{lemma:1-27-21}.
\end{proof}

The result follows by considering the Karush-Kuhn-Tucker conditions to incorporate the constraints.

\subsection{Proof of Theorem \ref{theorem:main}}\label{sec:algorithm_appendix:proof_Theorem_main}

Let $T_\omega \leq T$ be the stopping time of the algorithm. Starting from the definition of regret we get:

\begingroup
\allowdisplaybreaks
\begin{align*}
    regret(T) &= \OPT - \sum_{t=1}^T r_t(a_t)
    \\
    &= \OPT - \sum_{t=1}^{T_\omega} r_t(a_t)
    \\
    &= \OPT - \sum_{t=1}^{N_\mathcal{S} T_0} r_t(a_t) - \sum_{t = N_\mathcal{S} T_0 + 1}^{T_\omega}  r_t(a_t)
    \\
    &\leq \OPT - \sum_{t = N_\mathcal{S} T_0 + 1}^{T_\omega}  r_t(a_t)
\end{align*}
\endgroup

\noindent where the inequality follows since $r_t(a_t) \in [0, 1]$. Now, let

\begin{equation*}
    R(T) := O\Big( C  m \sqrt{T \log \big( dT m / \zeta \big) \log(T)} +  C \epsilon_c  m^\frac{3}{2} T \sqrt{\log(T)} \Big)
\end{equation*}

The proof now proceeds with first stating Lemma \ref{lemma:diff-reward-optimistic} that allows us to work with the optimistic estimates instead of the actual realizations of the reward and the consumption. 

\begin{lemma}
With probability at least $(1 - \zeta)^3$ we have:
\begin{itemize}
    \item[a)] $| \sum_{t=N_\mathcal{S} T_0 + 1}^{T_\omega} r_t(a_t) - \bm{x}_t(a_t)^\top \tilde{\bm{\mu}}_{a_t, t} | \leq R(T)$
    
    \item[b)] $ \| \sum_{t=N_\mathcal{S} T_0 + 1}^{T_\omega} \bm{v}_t(a_t) - \bm{x}_t(a_t)^\top \tilde{\bm{W}}_{a_t, t} \|_\infty \leq R(T)$
    
\end{itemize}

\label{lemma:diff-reward-optimistic}
\end{lemma}

\begin{proof}
Considering part $a)$, we start with the probability of the event of interest, and we utilize Lemma \ref{lemma:optimism} $c)$ in order to make the first $(1 - \zeta)$ term show up, and the Azuma-Hoeffding inequality for the remaining two $(1 - \zeta)$ terms.

\begingroup
\allowdisplaybreaks
\begin{align*}
    & \Pr \Big[ |\sum_{t=N_\mathcal{S}T_0 + 1}^{T_\omega} r_t(a_t) - \bm{x}_t(a_t)^\top \tilde{\bm{\mu}}_{a_t, t}| \leq R(T) \Big]
    \\
    &= \Pr \Big[ |\sum_{t=N_\mathcal{S}T_0 + 1}^{T_\omega} r_t(a_t) - \bm{x}_t(a_t)^\top (\tilde{\bm{\mu}}_{a_t, t} + \bm{\mu}_{\hat{c}(a_t)} - \bm{\mu}_{\hat{c}(a_t)})| \leq R(T) \Big]
    \\
    &\geq \Pr \Big[ |\sum_{t=N_\mathcal{S}T_0 + 1}^{T_\omega} r_t(a_t) - \bm{x}_t(a_t)^\top \bm{\mu}_{\hat{c}(a_t)}| + | \sum_{t=N_\mathcal{S}T_0 + 1}^{T_\omega} \bm{x}_t(a_t)^\top (\bm{\mu}_{\hat{c}(a_t)} - \tilde{\bm{\mu}}_{a_t, t}) | \leq R(T) \Big]
    \\
    &\geq \Pr \Big[ |\sum_{t=N_\mathcal{S}T_0 + 1}^{T_\omega} r_t(a_t) - \bm{x}_t(a_t)^\top \bm{\mu}_{\hat{c}(a_t)}| \leq \frac{R(T)}{2} \Big] \cdot \Pr \Big[ | \sum_{t=N_\mathcal{S}T_0 + 1}^{T_\omega} \bm{x}_t(a_t)^\top (\bm{\mu}_{\hat{c}(a_t)} - \tilde{\bm{\mu}}_{a_t, t}) | \leq \frac{R(T)}{2} \Big]
    \\
    &\geq \Pr \Big[ |\sum_{t=N_\mathcal{S}T_0 + 1}^{T_\omega} r_t(a_t) - \bm{x}_t(a_t)^\top \bm{\mu}_{\hat{c}(a_t)}| \leq \frac{R(T)}{2} \Big] \cdot (1 - \zeta)
    \\
    &= \Pr \Big[ |\sum_{t=N_\mathcal{S}T_0 + 1}^{T_\omega} r_t(a_t) - \bm{x}_t(a_t)^\top (\bm{\mu}_{\hat{c}(a_t)} + \bm{\mu}_{c(a_t)} - \bm{\mu}_{c(a_t)})| \leq \frac{R(T)}{2} \Big] \cdot (1 - \zeta)
    \\
    &\geq \Pr \Big[ |\sum_{t=N_\mathcal{S}T_0 + 1}^{T_\omega} r_t(a_t) - \bm{x}_t(a_t)^\top \bm{\mu}_{c(a_t)}| +|\sum_{t=N_\mathcal{S}T_0 + 1}^{T_\omega}  \bm{x}_t(a_t)^\top (\bm{\mu}_{c(a_t)} - \bm{\mu}_{\hat{c}(a_t)} )|  \leq \frac{R(T)}{2} \Big] \cdot (1 - \zeta)
    \\
    &\geq \Pr \Big[ |\sum_{t=N_\mathcal{S}T_0 + 1}^{T_\omega} r_t(a_t) - \bm{x}_t(a_t)^\top \bm{\mu}_{c(a_t)}| + \sum_{t=N_\mathcal{S}T_0 + 1}^{T_\omega} \mathbbm{1}[\hat{c}(a_t) \neq c(a_t)]  \leq \frac{R(T)}{2} \Big] \cdot (1 - \zeta)
    \\
    &\geq \Pr \Big[ |\sum_{t=N_\mathcal{S}T_0 + 1}^{T_\omega} r_t(a_t) - \bm{x}_t(a_t)^\top \bm{\mu}_{c(a_t)}| \leq \frac{R(T)}{4} \Big] \cdot  \Pr \Big[ \sum_{t=N_\mathcal{S}T_0 + 1}^{T_\omega} \mathbbm{1}[\hat{c}(a_t) \neq c(a_t)]  \leq \frac{R(T)}{4} \Big] \cdot (1 - \zeta)
    \\
    &\geq \Pr \Big[ |\sum_{t=N_\mathcal{S}T_0 + 1}^{T_\omega} r_i(a_t) - \bm{x}_t(a_t)^\top \bm{\mu}_{c(a_t)}| \leq \frac{R(T)}{4} \Big] \cdot  \Big( 1 - 2 \exp \Big( - \frac{(R(T)/4)^2}{2(T_\omega - N_{\mathcal{S}} T_0)} \Big) \Big) \cdot (1 - \zeta)
    \\
    &\geq \Pr \Big[ |\sum_{t=N_\mathcal{S}T_0 + 1}^{T_\omega} r_t(a_t) - \bm{x}_t(a_t)^\top \bm{\mu}_{c(a_t)}| \leq \frac{R(T)}{4} \Big] \cdot (1 - \zeta)^2
    \\
    &\geq \Big( 1 - 2 \exp \Big( - \frac{(R(T)/4)^2}{2(T_\omega - N_{\mathcal{S}} T_0)} \Big) \Big) \cdot (1 - \zeta)^2
    \\
    &\geq (1 - \zeta)^3
\end{align*}
\endgroup

The proof for part $b)$ follows the same steps.
\end{proof}

Now, let $\mathcal{S}_1$ denote the subset of $\mathcal{S}$ that contains correctly clustered arms,

\begin{equation*}
    \mathcal{S}_1 := \{ a \in \mathcal{S}: \hat{c}(a) = c(a) \}
\end{equation*}

The following lemma provides a lower bound related to the choice of the algorithm.

\begin{lemma}\label{eq:1-15-a}
For $t > N_\mathcal{S} T_0$, the following inequality holds with high probability:

\begin{equation*}
    \bm{x}_t(a_t)^\top ( \tilde{\bm{\mu}}_{a_t, t} -  Z \tilde{\bm{W}}_{a_t, t} \bm{\theta}_t) \geq \frac{1}{\sum_{a^\prime \in  \mathcal{S}_1} \pi^*(a^\prime, \bm{X}_t)} \sum_{a \in \mathcal{S}_1} \pi^*(a, \bm{X}_t) \cdot  \bm{x}_t(a)^\top  \big( \bm{\mu}_{c(a)} - Z W_{c(a)} \bm{\theta}_t \big)  
\end{equation*}
\end{lemma}
\begin{proof}
Let $\Pi^\mathcal{S}$ denote the set of static policies that assign non-zero probability only to arms in $\mathcal{S}$, and let $\Pi^{\mathcal{S}_1}$ be defined equivalently. Then, for $t > N_\mathcal{S}  T_0$ we have

\begin{align*}
     & \bm{x}_t(a_t)^\top ( \tilde{\bm{\mu}}_{a_t, t} -  Z \tilde{\bm{W}}_{a_t, t} \bm{\theta}_t) \\
     &\geq \max_{\pi^\mathcal{S} \in \Pi^\mathcal{S}} \sum_{a \in \mathcal{S}} \pi^\mathcal{S}(a, \bm{X}_t) \cdot  \bm{x}_t(a)^\top ( \tilde{\bm{\mu}}_{a, t} -  Z \tilde{\bm{W}}_{a, t} \bm{\theta}_t)
     \\
     & \geq \max_{\pi^\mathcal{S} \in \Pi^{\mathcal{S}_1}} \sum_{a \in \mathcal{S}_1} \pi^\mathcal{S}(a, \bm{X}_t) \cdot  \bm{x}_t(a)^\top ( \tilde{\bm{\mu}}_{a, t} -  Z \tilde{\bm{W}}_{a, t} \bm{\theta}_t)
     \\
     & \geq \frac{1}{\sum_{a^\prime \in \mathcal{S}_1} \pi^*(a^\prime, \bm{X}_t)} \sum_{a \in \mathcal{S}_1} \pi^*(a, \bm{X}_t) \cdot  \bm{x}_t(a)^\top ( \tilde{\bm{\mu}}_{a, t} -  Z \tilde{\bm{W}}_{a, t} \bm{\theta}_t)
     \\
     & \geq \frac{1}{\sum_{a^\prime \in \mathcal{S}_1} \pi^*(a^\prime, \bm{X}_t)} \sum_{a \in \mathcal{S}_1} \pi^*(a, \bm{X}_t) \cdot  \bm{x}_t(a)^\top (  \bm{\mu}_{\hat{c}(a)} -  Z  \bm{W}_{\hat{c}(a)} \bm{\theta}_t)
     \\
     & = \frac{1}{\sum_{a^\prime \in \mathcal{S}_1} \pi^*(a^\prime, \bm{X}_t)} \sum_{a \in \mathcal{S}_1} \pi^*(a, \bm{X}_t) \cdot  \bm{x}_t(a)^\top (  \bm{\mu}_{c(a)} -  Z  \bm{W}_{c(a)} \bm{\theta}_t)
\end{align*}

\noindent where the first inequality follows from the choice of the algorithm, the second from restricting the set of arms to $\mathcal{S}_1$, the third from considering the normalization of $\pi^*$ as a policy in $\Pi^{\mathcal{S}_1}$, the fourth from Lemma \ref{lemma:optimism} $a)$ and $b)$, and the last equality from the definition of $\mathcal{S}_1$.
\end{proof}

\begin{lemma}
\label{lemma:1-16-A}
The following inequality holds with high probability:

\begin{equation*}
    \sum_{t=N_\mathcal{S} T_0 + 1}^{T_\omega} \E_{\bm{X}_t} \Big[ \bm{x}_t(a_t)^\top \tilde{\bm{\mu}}_{a_t, t} \Big] \geq O\Big(   \frac{N_\mathcal{S} \cdot T_\omega}{K \cdot T} \OPT  +  Z \sum_{t=N_\mathcal{S} T_0 + 1}^{T_\omega} \E_{\bm{X}_t} \Big[  \bm{x}_t(a_t)^\top  \tilde{\bm{W}}_{a_t, t} -  \frac{N_\mathcal{S}}{K} \cdot \frac{B}{T} \bm{1}_d \Big]  \bm{\theta}_t  \Big)
\end{equation*}
\end{lemma}
\begin{proof}
Lemma \ref{eq:1-15-a} implies the weaker condition where expectation is taken over $\bm{X_t}$ only and it is conditional on the past realizations of the context:

\begin{align*}
    & \E_{\bm{X}_t} \Big[ \bm{x}_t(a_t)^\top ( \tilde{\bm{\mu}}_{a_t, t} -  Z \tilde{\bm{W}}_{a_t, t} \bm{\theta}_t) \Big] 
    \\
    &\geq \E_{\bm{X}_t} \Big[  \frac{1}{\sum_{a^\prime \in  \mathcal{S}_1} \pi^*(a^\prime, \bm{X}_t)}  \sum_{ a \in \mathcal{S}_1} \pi^*(a, \bm{X}_t) \cdot  \bm{x}_t(a)^\top  \big( \bm{\mu}_{c(a)} - Z W_{c(a)} \bm{\theta}_t \big)   \Big]
    \\
    &\geq \E_{\bm{X}_t} \Big[  \sum_{ a \in \mathcal{S}_1} \pi^*(a, \bm{X}_t) \cdot  \bm{x}_t(a)^\top  \big( \bm{\mu}_{c(a)} - Z W_{c(a)} \bm{\theta}_t \big)   \Big]
    \\
    &= \E_{\bm{X}_t} \Big[  \sum_{ a \in \mathcal{S}_1} \pi^*(a, \bm{X}_t) \cdot  \bm{x}_t(a)^\top   \bm{\mu}_{c(a)} \Big] - Z \E_{\bm{X}_t} \Big[ \sum_{ a \in \mathcal{S}_1} \pi^*(a, \bm{X}_t) \cdot  \bm{x}_t(a)^\top \bm{W}_{c(a)}  \Big] \bm{\theta}_t
    \\
    &= O\Big( \frac{N_\mathcal{S}}{K} \Big)  \cdot \Big( \frac{\OPT}{T}  - Z \frac{B}{T} \bm{1}_d \bm{\theta}_t \Big)
\end{align*}
where the second step holds since the probability-normalization term is greater than one. From Lemma \ref{lemma:su2016-1} the size of $\mathcal{S}_1$ is $N_\mathcal{S}(1 - o(N_\mathcal{S}^{-1})) = O(N_\mathcal{S})$. We get the statement of the lemma by summing from period $N_\mathcal{S}T_0+1$ to period $T_\omega$.
\end{proof}

Since in the first $N_\mathcal{S} T_0$ periods the choices are made randomly, and so $N_\mathcal{S} T_0$ of the budget can potentially be consumed, the following lemma which is known from the literature is expressed in terms of $B^\prime = B - N_\mathcal{S} T_0$ and $ T^\prime = T - N_\mathcal{S} T_0$ rather than $B$ and $T$.

\begin{lemma}[Lemma 9 of \cite{agrawal2016a}]
\begin{equation*}
    \sum_{t=N_\mathcal{S}T_0+1}^{T_\omega} \Big( \bm{x}_t(a_t)^\top \tilde{\bm{W}}_{a_t, t} - \frac{B^\prime}{T^\prime} \bm{1}_d \Big) \bm{\theta}_t \geq B^\prime \Big( 1 - \frac{T_\omega - N_\mathcal{S} T_0}{T^\prime} \Big) - R(T)
\end{equation*}
\label{lemma:1-16-B}
\end{lemma}

\begin{lemma}
\label{lemma:1-23-A}
$\frac{N_\mathcal{S} \cdot B}{K \cdot T} < 2 \frac{B^\prime}{T^\prime}$
\end{lemma}
\begin{proof}
$
    \frac{N_\mathcal{S} B T^\prime}{K T B^\prime} \leq  \frac{N_\mathcal{S} B }{K B^\prime} \leq \frac{B}{B^\prime} = \frac{1}{1 - \frac{N_\mathcal{S} T_0}{B}} < 2
$, since $B/2 > N_\mathcal{S} T_0$.
\end{proof}

Thus, we have
\begin{align}
     & ~ \sum_{t=N_\mathcal{S} T_0 + 1}^{T_\omega} \E_{\bm{X}_t} \Big[ \bm{x}_t(a_t)^\top \tilde{\bm{\mu}}_{a_t, t} \Big]  \notag \\
     &\geq O\Big(   \frac{N_\mathcal{S}  T_\omega}{K  T} \OPT  +  Z \sum_{t=N_\mathcal{S} T_0 + 1}^{T_\omega} \E_{\bm{X}_t} \Big[  \bm{x}_t(a_t)^\top  \tilde{\bm{W}}_{a_t, t} - 2 \frac{B^\prime}{T^\prime} \bm{1}_d \Big]  \bm{\theta}_t  \Big)
     \label{eq:1-25-B}
     \\
     &\geq O\Big(   \frac{N_\mathcal{S}  T_\omega}{K  T} \OPT  +  Z \Big( 2 B^\prime \big( 1 - \frac{T_\omega - N_\mathcal{S} T_0}{T^\prime}  \big) - R(T) \Big) \Big)
     \notag \\
     &\geq O\Big(   \frac{N_\mathcal{S}  T_\omega}{K  T} \OPT  +  \frac{N_\mathcal{S} \OPT }{2 K B^\prime } \Big( 2 B^\prime \big( 1 - \frac{T_\omega - N_\mathcal{S} T_0}{T^\prime}  \big) - R(T) \Big) \Big)
     \notag \\
     &\geq O\Big( \frac{N_\mathcal{S}}{K} \OPT \Big( \frac{T_\omega}{T} + 1 - \frac{T_\omega - N_\mathcal{S} T_0}{T^\prime} - \frac{R(T)}{2B^\prime} \Big) \Big)
     \notag \\
     &\geq O\Big( \frac{N_\mathcal{S}}{K} \OPT \Big( 1 - \frac{R(T)}{B^\prime} \Big) \Big)
     \label{eq:1-17-A}
\end{align}
where the first step follows from Lemmas \ref{lemma:1-16-A} and \ref{lemma:1-23-A}, the second from Lemma \ref{lemma:1-16-B},  and the third from Lemma \ref{lemma:Z}.

From Lemma \ref{lemma:diff-reward-optimistic}, the bound in Eq. (\ref{eq:1-17-A}) applies to $\sum_{t=N_\mathcal{S} T_0 + 1}^{T_\omega} \E_{\bm{X}_t} \big[ r_t(a_t) \big]$ by adding $R(T)$, and an application of the Azuma-Hoeffding inequality to the realized reward $\sum_{t=N_\mathcal{S} T_0 + 1}^{T_\omega}  r_t(a_t)$ then gives:
\begin{align*}
    regret(T) &\leq O \Big( \OPT - \frac{N_\mathcal{S}}{K}\OPT \big( 1 - \frac{R(T)}{ B^\prime} \big) + R(T) \Big)
    \\
    &= O \Big( \OPT \big( 1 -  \frac{N_\mathcal{S}}{K} \big)  + R(T) \big( 1 + \frac{N_\mathcal{S} \OPT}{ K B^\prime} \big) \Big).
\end{align*}


%% file: main.bbl
\newcommand{\etalchar}[1]{$^{#1}$}
\begin{thebibliography}{BBMR18}

\bibitem[AAKS13]{abraham2013}
Ittai Abraham, Omar Alonso, Vasilis Kandylas, and Aleksandrs Slivkins.
\newblock Adaptive crowdsourcing algorithms for the bandit survey problem.
\newblock In {\em Conference on learning theory}, pages 882--910. PMLR, 2013.

\bibitem[AAT19]{amani2019}
Sanae Amani, Mahnoosh Alizadeh, and Christos Thrampoulidis.
\newblock Linear stochastic bandits under safety constraints.
\newblock {\em Advances in Neural Information Processing Systems},
  32:9256--9266, 2019.

\bibitem[AB16]{ando2016}
Tomohiro Ando and Jushan Bai.
\newblock Panel data models with grouped factor structure under unknown group
  membership.
\newblock {\em Journal of Applied Econometrics}, 31(1):163--191, 2016.

\bibitem[ACBF02]{auer2002b}
Peter Auer, Nicolo Cesa-Bianchi, and Paul Fischer.
\newblock Finite-time analysis of the multiarmed bandit problem.
\newblock {\em Machine learning}, 47(2):235--256, 2002.

\bibitem[AD14]{agrawal2014}
Shipra Agrawal and Nikhil~R Devanur.
\newblock Bandits with concave rewards and convex knapsacks.
\newblock In {\em Proceedings of the fifteenth ACM conference on Economics and
  computation}, pages 989--1006, 2014.

\bibitem[AD16]{agrawal2016a}
Shipra Agrawal and Nikhil Devanur.
\newblock Linear contextual bandits with knapsacks.
\newblock {\em Advances in Neural Information Processing Systems},
  29:3450--3458, 2016.

\bibitem[ADL16]{agrawal2016b}
Shipra Agrawal, Nikhil~R Devanur, and Lihong Li.
\newblock An efficient algorithm for contextual bandits with knapsacks, and an
  extension to concave objectives.
\newblock In {\em Conference on Learning Theory}, pages 4--18. PMLR, 2016.

\bibitem[AHK{\etalchar{+}}14]{agarwal2014}
Alekh Agarwal, Daniel Hsu, Satyen Kale, John Langford, Lihong Li, and Robert
  Schapire.
\newblock Taming the monster: A fast and simple algorithm for contextual
  bandits.
\newblock In {\em International Conference on Machine Learning}, pages
  1638--1646. PMLR, 2014.

\bibitem[Aue02]{auer2002a}
Peter Auer.
\newblock Using confidence bounds for exploitation-exploration trade-offs.
\newblock {\em Journal of Machine Learning Research}, 3(Nov):397--422, 2002.

\bibitem[AYPS11]{abbasi2011}
Yasin Abbasi-Yadkori, D{\'a}vid P{\'a}l, and Csaba Szepesv{\'a}ri.
\newblock Improved algorithms for linear stochastic bandits.
\newblock {\em Advances in neural information processing systems},
  24:2312--2320, 2011.

\bibitem[BBMR18]{balakrishnan2018}
Avinash Balakrishnan, Djallel Bouneffouf, Nicholas Mattei, and Francesca Rossi.
\newblock Using contextual bandits with behavioral constraints for constrained
  online movie recommendation.
\newblock In {\em IJCAI}, pages 5802--5804, 2018.

\bibitem[BH21]{ban2021}
Yikun Ban and Jingrui He.
\newblock Local clustering in contextual multi-armed bandits.
\newblock In {\em Proceedings of the Web Conference 2021}, pages 2335--2346,
  2021.

\bibitem[BKS18]{badanidiyuru2018}
Ashwinkumar Badanidiyuru, Robert Kleinberg, and Aleksandrs Slivkins.
\newblock Bandits with knapsacks.
\newblock {\em Journal of the ACM (JACM)}, 65(3):1--55, 2018.

\bibitem[BLS14]{badanidiyuru2014}
Ashwinkumar Badanidiyuru, John Langford, and Aleksandrs Slivkins.
\newblock Resourceful contextual bandits.
\newblock In {\em Conference on Learning Theory}, pages 1109--1134. PMLR, 2014.

\bibitem[CDJ21]{carlsson2021}
Emil Carlsson, Devdatt Dubhashi, and Fredrik~D Johansson.
\newblock Thompson sampling for bandits with clustered arms.
\newblock {\em arXiv preprint arXiv:2109.01656}, 2021.

\bibitem[CES20]{cayci2020}
Semih Cayci, Atilla Eryilmaz, and R~Srikant.
\newblock Budget-constrained bandits over general cost and reward
  distributions.
\newblock In {\em International Conference on Artificial Intelligence and
  Statistics}, pages 4388--4398. PMLR, 2020.

\bibitem[CLRS11]{chu2011}
Wei Chu, Lihong Li, Lev Reyzin, and Robert Schapire.
\newblock Contextual bandits with linear payoff functions.
\newblock In {\em Proceedings of the Fourteenth International Conference on
  Artificial Intelligence and Statistics}, pages 208--214. JMLR Workshop and
  Conference Proceedings, 2011.

\bibitem[CLS20]{chen2020}
Sitan Chen, Jerry Li, and Zhao Song.
\newblock Learning mixtures of linear regressions in subexponential time via
  fourier moments.
\newblock In {\em Proceedings of the 52nd Annual ACM SIGACT Symposium on Theory
  of Computing}, pages 587--600, 2020.

\bibitem[CMPC21]{chen2021}
Yanxi Chen, Cong Ma, H~Vincent Poor, and Yuxin Chena.
\newblock Learning mixtures of low-rank models.
\newblock {\em IEEE Transactions on Information Theory}, 2021.

\bibitem[DHK08]{dani2008}
Varsha Dani, Thomas~P Hayes, and Sham~M Kakade.
\newblock Stochastic linear optimization under bandit feedback.
\newblock In {\em Conference on Learning Theory}, 2008.

\bibitem[DJSW11]{devanur2011}
Nikhil~R Devanur, Kamal Jain, Balasubramanian Sivan, and Christopher~A Wilkens.
\newblock Near optimal online algorithms and fast approximation algorithms for
  resource allocation problems.
\newblock In {\em Proceedings of the 12th ACM conference on Electronic
  commerce}, pages 29--38, 2011.

\bibitem[FGT92]{flajolet1992}
Philippe Flajolet, Daniele Gardy, and Lo{\"y}s Thimonier.
\newblock Birthday paradox, coupon collectors, caching algorithms and
  self-organizing search.
\newblock {\em Discrete Applied Mathematics}, 39(3):207--229, 1992.

\bibitem[FHK{\etalchar{+}}10]{feldman2010}
Jon Feldman, Monika Henzinger, Nitish Korula, Vahab~S Mirrokni, and Cliff
  Stein.
\newblock Online stochastic packing applied to display ad allocation.
\newblock In {\em European Symposium on Algorithms}, pages 182--194. Springer,
  2010.

\bibitem[Ful09]{fuller2009}
Wayne~A Fuller.
\newblock {\em Measurement error models}, volume 305.
\newblock John Wiley \& Sons, 2009.

\bibitem[GLK{\etalchar{+}}17]{gentile2017}
Claudio Gentile, Shuai Li, Purushottam Kar, Alexandros Karatzoglou, Giovanni
  Zappella, and Evans Etrue.
\newblock On context-dependent clustering of bandits.
\newblock In {\em International Conference on Machine Learning}, pages
  1253--1262. PMLR, 2017.

\bibitem[GLZ14]{gentile2014}
Claudio Gentile, Shuai Li, and Giovanni Zappella.
\newblock Online clustering of bandits.
\newblock In {\em International Conference on Machine Learning}, pages
  757--765. PMLR, 2014.

\bibitem[GV19]{gu2019}
Jiaying Gu and Stanislav Volgushev.
\newblock Panel data quantile regression with grouped fixed effects.
\newblock {\em Journal of Econometrics}, 213(1):68--91, 2019.

\bibitem[ISSS19]{immorlica2019}
Nicole Immorlica, Karthik~Abinav Sankararaman, Robert Schapire, and Aleksandrs
  Slivkins.
\newblock Adversarial bandits with knapsacks.
\newblock In {\em 2019 IEEE 60th Annual Symposium on Foundations of Computer
  Science (FOCS)}, pages 202--219. IEEE, 2019.

\bibitem[KS20]{kesselheim2020}
Thomas Kesselheim and Sahil Singla.
\newblock Online learning with vector costs and bandits with knapsacks.
\newblock In {\em Conference on Learning Theory}, pages 2286--2305. PMLR, 2020.

\bibitem[KSS{\etalchar{+}}20]{kong2020}
Weihao Kong, Raghav Somani, Zhao Song, Sham Kakade, and Sewoong Oh.
\newblock Meta-learning for mixed linear regression.
\newblock In {\em International Conference on Machine Learning}, pages
  5394--5404. PMLR, 2020.

\bibitem[LL18]{li2018b}
Yuanzhi Li and Yingyu Liang.
\newblock Learning mixtures of linear regressions with nearly optimal
  complexity.
\newblock In {\em Conference On Learning Theory}, pages 1125--1144. PMLR, 2018.

\bibitem[LN12]{lin2012}
Chang-Ching Lin and Serena Ng.
\newblock Estimation of panel data models with parameter heterogeneity when
  group membership is unknown.
\newblock {\em Journal of Econometric Methods}, 1(1):42--55, 2012.

\bibitem[LWW21]{li2021}
Chuanhao Li, Qingyun Wu, and Hongning Wang.
\newblock Unifying clustered and non-stationary bandits.
\newblock In {\em International Conference on Artificial Intelligence and
  Statistics}, pages 1063--1071. PMLR, 2021.

\bibitem[LZ07]{langford2007}
John Langford and Tong Zhang.
\newblock Epoch-greedy algorithm for multi-armed bandits with side information.
\newblock {\em Advances in Neural Information Processing Systems (NIPS 2007)},
  20:1, 2007.

\bibitem[LZ18]{li2018a}
Shuai Li and Shengyu Zhang.
\newblock Online clustering of contextual cascading bandits.
\newblock In {\em Thirty-Second AAAI Conference on Artificial Intelligence},
  2018.

\bibitem[NL14]{nguyen2014}
Trong~T Nguyen and Hady~W Lauw.
\newblock Dynamic clustering of contextual multi-armed bandits.
\newblock In {\em Proceedings of the 23rd ACM International Conference on
  Conference on Information and Knowledge Management}, pages 1959--1962, 2014.

\bibitem[OW21]{okui2021}
Ryo Okui and Wendun Wang.
\newblock Heterogeneous structural breaks in panel data models.
\newblock {\em Journal of Econometrics}, 220(2):447--473, 2021.

\bibitem[RFTT19]{rangi2019}
Anshuka Rangi, Massimo Franceschetti, and Long Tran-Thanh.
\newblock Unifying the stochastic and the adversarial bandits with knapsack.
\newblock In {\em IJCAI}, 2019.

\bibitem[SB18]{sutton2018}
Richard~S Sutton and Andrew~G Barto.
\newblock {\em Reinforcement learning: An introduction}.
\newblock MIT press, 2018.

\bibitem[SBF17]{schwartz2017}
Eric~M Schwartz, Eric~T Bradlow, and Peter~S Fader.
\newblock Customer acquisition via display advertising using multi-armed bandit
  experiments.
\newblock {\em Marketing Science}, 36(4):500--522, 2017.

\bibitem[Sli11]{slivkins2011}
Aleksandrs Slivkins.
\newblock Contextual bandits with similarity information.
\newblock In {\em Proceedings of the 24th annual Conference On Learning
  Theory}, pages 679--702. JMLR Workshop and Conference Proceedings, 2011.

\bibitem[SS{\etalchar{+}}11]{shalev2011}
Shai Shalev-Shwartz et~al.
\newblock Online learning and online convex optimization.
\newblock {\em Foundations and trends in Machine Learning}, 4(2):107--194,
  2011.

\bibitem[SSP16]{su2016}
Liangjun Su, Zhentao Shi, and Peter~CB Phillips.
\newblock Identifying latent structures in panel data.
\newblock {\em Econometrica}, 84(6):2215--2264, 2016.

\bibitem[SST11]{srebro2011}
Nati Srebro, Karthik Sridharan, and Ambuj Tewari.
\newblock On the universality of online mirror descent.
\newblock In {\em Advances in neural information processing systems}, pages
  2645--2653, 2011.

\bibitem[SWJ19]{su2019}
Liangjun Su, Xia Wang, and Sainan Jin.
\newblock Sieve estimation of time-varying panel data models with latent
  structures.
\newblock {\em Journal of Business \& Economic Statistics}, 37(2):334--349,
  2019.

\bibitem[TJLL14]{tang2014}
Liang Tang, Yexi Jiang, Lei Li, and Tao Li.
\newblock Ensemble contextual bandits for personalized recommendation.
\newblock In {\em Proceedings of the 8th ACM Conference on Recommender
  Systems}, pages 73--80, 2014.

\bibitem[TM17]{tewari2017}
Ambuj Tewari and Susan~A Murphy.
\newblock From ads to interventions: Contextual bandits in mobile health.
\newblock In {\em Mobile Health}, pages 495--517. Springer, 2017.

\bibitem[TRSA13]{tang2013}
Liang Tang, Romer Rosales, Ajit Singh, and Deepak Agarwal.
\newblock Automatic ad format selection via contextual bandits.
\newblock In {\em Proceedings of the 22nd ACM international conference on
  Information \& Knowledge Management}, pages 1587--1594, 2013.

\bibitem[WM01]{wansbeek2001}
Tom Wansbeek and Erik Meijer.
\newblock Measurement error and latent variables.
\newblock {\em A companion to theoretical econometrics}, pages 162--179, 2001.

\bibitem[WSLJ15]{wu2015}
Huasen Wu, R~Srikant, Xin Liu, and Chong Jiang.
\newblock Algorithms with logarithmic or sublinear regret for constrained
  contextual bandits.
\newblock {\em Advances in Neural Information Processing Systems}, 28:433--441,
  2015.

\bibitem[YLQY20]{yang2020}
Mengyue Yang, Qingyang Li, Zhiwei Qin, and Jieping Ye.
\newblock Hierarchical adaptive contextual bandits for resource constraint
  based recommendation.
\newblock In {\em Proceedings of The Web Conference 2020}, pages 292--302,
  2020.

\bibitem[ZJD16]{zhong2016}
Kai Zhong, Prateek Jain, and Inderjit~S Dhillon.
\newblock Mixed linear regression with multiple components.
\newblock In {\em NIPS}, pages 2190--2198, 2016.

\end{thebibliography}
